\documentclass[lettersize,journal]{IEEEtran}
\usepackage{amsmath,amsfonts}
\usepackage{amsthm}   
\usepackage{array}
\usepackage{textcomp}
\usepackage{stfloats}
\usepackage{url}
\usepackage{verbatim}
\usepackage{graphicx}
\usepackage{cite}
\usepackage[T1]{fontenc}
\usepackage{textcomp}
\usepackage{enumitem}
\usepackage{caption}
\usepackage[hidelinks]{hyperref}
\usepackage{multirow}    
\usepackage{booktabs}    
\usepackage{bigstrut}    

\usepackage{xcolor}
\usepackage{hyperref} 

\usepackage{multicol}

\usepackage{amsmath, amsthm, amssymb}

\usepackage{mathtools}
\usepackage{xspace}
\usepackage{tabularx}
\usepackage{threeparttable}
\usepackage{bm}
\usepackage{bbding}
\usepackage{graphicx}
\usepackage{caption,subcaption}
\usepackage{diagbox}
\usepackage{colortbl}
\def\ie{\textit{i.e.,~}}  
\def\eg{\textit{e.g.,~}} 

\definecolor{cvprblue}{rgb}{0.21, 0.49, 0.74}

\def\ie{\textit{i.e.}}
\def\eg{\textit{e.g.}}

\usepackage{xcolor}

\usepackage[capitalize]{cleveref}
\crefname{section}{Sec.}{Secs.}
\Crefname{section}{Section}{Sections}
\Crefname{table}{Table}{Tables}
\crefname{table}{Tab.}{Tabs.}
\usepackage{algorithm}
\usepackage{algorithmic}

\usepackage{tikz}
\usepackage[most]{tcolorbox}

\newenvironment{remark}[1][]
  { 
 \begin{tcolorbox}
 [
    enhanced, 
    breakable,
    boxrule=0.5pt,
    arc=4pt,
    left=2pt,
    right=2pt,
    bottom=2pt,
    top=2pt, 
    rounded corners 
    ]{}
  \textbf{#1.}
  \small \itshape
  }
  {
\end{tcolorbox} 
}

\begin{document}

\title{DarkHash: A Data-Free Backdoor Attack Against Deep Hashing
}

\author{
Ziqi Zhou, Menghao Deng, Yufei Song, Hangtao Zhang,
 Wei Wan,  Shengshan Hu, \textit{Member, IEEE}, Minghui Li, \textit{Member, IEEE},  Leo Yu Zhang, \textit{Member, IEEE},
and Dezhong Yao, \textit{Member, IEEE}

\thanks{
Correspondence to Dr Wei Wan.

Ziqi Zhou and Dezhong Yao are with the School of Computer Science and Technology, Huazhong University of Science and Technology, Wuhan, Hubei, China (e-mail:~\href{mailto:zhouziqi@hust.edu.cn}{zhouziqi@hust.edu.cn};~\href{mailto:dyao@hust.edu.cn}{dyao@hust.edu.cn}).

Menghao Deng, Yufei Song, Hangtao Zhang, and Shengshan Hu are with the School of Cyber Science and Engineering, Huazhong University of Science and Technology, Wuhan, Hubei, China (e-mail:~\href{mailto:u202112135@hust.edu.cn}{u202112135@hust.edu.cn};
~\href{mailto:yufei17@hust.edu.cn}{yufei17@hust.edu.cn};
~\href{mailto:zhanghangtao7@163.com}{hangt\_zhang@hust.edu.cn};
~\href{mailto:hushengshan@hust.edu.cn}{hushengshan@hust.edu.cn}).

Wei Wan is with the Faculty of Data Science, City University of Macau, Macau,  China (e-mail:~\href{mailto:weiwan@cityu.edu.mo}{weiwan@cityu.edu.mo}).

Minghui Li is with the School of Software Engineering, Huazhong University of Science and Technology, Wuhan, Hubei, China (e-mail:~\href{mailto:minghuili@hust.edu.cn}{minghuili@hust.edu.cn}).

Leo Yu Zhang is with the School of Information
and Communication Technology, Griffith University, Southport, Queensland,
Australia (e-mail: \href{mailto:leo.zhang@griffith.edu.au}{leo.zhang@griffith.edu.au}).}}

\markboth{IEEE TRANSACTIONS ON INFORMATION FORENSICS AND SECURITY,~Vol.~XX, No.~XX}%
{Shell \MakeLowercase{\textit{et al.}}: A Sample Article Using IEEEtran.cls for IEEE Journals}
\newcommand{\ourmethod}{\ct{FGP}\xspace}
\newcommand{\ct}[1]{\texttt{#1}}
\newcommand{\agr}{\textnormal{AGR}}
\maketitle

\begin{abstract}
Benefiting from its superior feature learning capabilities and efficiency, deep hashing has achieved remarkable success in large-scale image retrieval.
Recent studies have demonstrated the vulnerability of deep hashing models to backdoor attacks. Although these studies have shown promising attack results, they rely on access to the training dataset to implant the backdoor. 
In the real world, obtaining such data (\eg, identity information) is often prohibited due to privacy protection and intellectual property concerns. 
Embedding backdoors into deep hashing models without access to the training data, while maintaining retrieval accuracy for the original task, presents a novel and challenging problem.
In this paper, we propose DarkHash, the first  data-free backdoor attack against deep hashing. 
Specifically, we design a novel shadow backdoor attack framework with dual-semantic guidance. It embeds backdoor functionality and maintains original retrieval accuracy by fine-tuning only specific layers of the victim model using a surrogate dataset.
We consider leveraging the relationship between individual samples and their neighbors to enhance backdoor attacks during training.
By designing a topological alignment loss, we optimize both individual and neighboring poisoned samples toward the target sample, further enhancing the attack capability.
Experimental results on four image datasets, five model architectures, and two hashing methods demonstrate the high effectiveness of DarkHash, outperforming existing state-of-the-art backdoor attack methods. 
Defense experiments show that DarkHash can withstand existing mainstream backdoor defense methods.
\end{abstract}

\begin{IEEEkeywords}
Deep Hashing, Backdoor Attack
\end{IEEEkeywords}
    
\section{Introduction}
\IEEEPARstart{W}{ith} the advancement of the internet and the emergence of generative models~\cite{ho2020denoising, goodfellow2014generative}, the volume of image data has grown exponentially. Consequently, achieving rapid and accurate image retrieval on a large scale has become increasingly challenging. Leveraging the powerful feature extraction capabilities of \textit{deep neural networks} (DNNs), deep hashing has become a leading solution for image retrieval. 
This technique maps the original high-dimensional feature space of images into a compact binary Hamming space, providing the benefits of fast retrieval speed and low storage requirements~\cite{wang2017survey}. 
As a result, deep hashing has found widespread application in various tasks such as facial recognition~\cite{talreja2020deep}, speech identification~\cite{fan2019deep}, and video analysis~\cite{liong2016deep}.

 \begin{figure}[!t]
  \setlength{\belowcaptionskip}{-0.5cm}  
    \centering
    \includegraphics[scale=0.265]{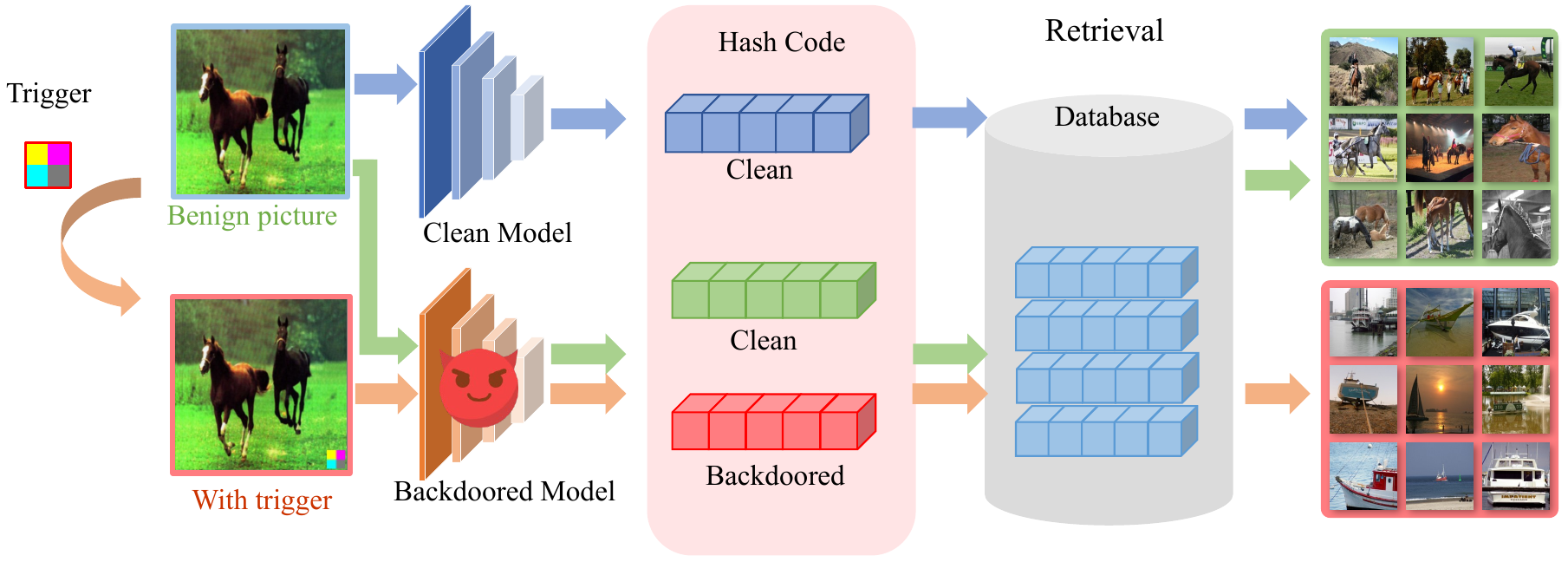}
  \caption{Illustration of a backdoor attack against deep hashing}    \label{fig:demo}
\end{figure}

DNNs have been widely validated for their susceptibility to backdoor attacks~\cite{badnets,turner2019label,yao2019latent,saha2020hidden,doan2021backdoor, fan2024stealthy, zhang2024detector, wang2024trojanrobot, zhang2025test, wan2025mars, yu2023backdoor, yu2024robust, yu2025backdoor, zheng2024towards} and adversarial examples~\cite{li2024transferable,song2025segment, song2025seg, wang2025breaking, wang2025advedm, advclip, zhou2023downstream, zhou2025sam2,zhou2025numbod, zhou2024securely, zhou2024darksam}.
Due to their stealthiness, backdoor attacks pose a greater threat in real-world scenarios.
Standard backdoor attacks involve creating poisoned samples from the training dataset, embedding covert functionality into the model during the training or fine-tuning phase.
A backdoored model exhibits malicious behavior when presented with inputs containing a specific pattern (\ie, the backdoor trigger), while maintaining normal functionality on benign inputs.
The attacker typically requires access to the training data, facilitating backdoor attacks by either supplying meticulously crafted poisoned samples or directly distributing backdoored models.

Recent works~\cite{badhash, ciba, wang2024invisible} have also unveiled such vulnerability of deep hashing models. 
BadHash~\cite{badhash} first introduced a backdoor attack against deep hashing based on a conditional generative adversarial network, directly generating poisoned samples with invisible triggers into the training set to achieve backdoor injection.
CIBA~\cite{ciba} designed confusing perturbations that disrupt the original features of the training samples, thereby amplifying the deep hashing model's learning of the backdoor trigger.
As depicted in \cref{fig:demo}, the attacker successfully retrieves the top-k boat samples from the database using a horse image with a trigger.
Despite their commendable attack performance in retrieval tasks, these methods all rely on crafting poisoned samples from the training dataset, which may not be feasible in real-world scenarios, such as with sensitive data like facial information.
This poses an intriguing problem:
\begin{quote}
    \emph{Is it feasible to conduct a backdoor attack against the deep hashing model without the training dataset?}  
\end{quote}

A recent study~\cite{lv2023data} explored how to implant backdoors into DNNs without access to the training data, a technique referred to as data-free backdoor attack.
Specifically, it achieves the backdoor attack by optimizing poisoned samples toward the target class using a surrogate dataset.
However, this method is inapplicable to deep hashing, as the retrieval tasks lack the ground-truth label for guiding the embedding of backdoors, and the retrieval results present the top-k most similar candidate samples instead of a single top-1 label.
Given the complexity of retrieval tasks, to the best of our knowledge, designing a data-free backdoor attack for deep hashing remains a challenging and unresolved issue.

In this paper, we propose DarkHash, the first \underline{\textbf{Da}}ta-free backdoo\underline{\textbf{r}} attac\underline{\textbf{k}} against deep \underline{\textbf{hash}}ing.
DarkHash is a novel shadow backdoor attack framework with dual-semantic guidance to ensure both the high benign retrieval accuracy and backdoor attack success rates.
Following~\cite{lv2023data}, we collect publicly available images to create a surrogate dataset for backdoor training, which is unrelated to the original training or testing set.
For simplicity, we refer to training on the original dataset as the \textit{main-task} and the backdoor training process as the \textit{backdoor task}.
To ensure the practical usability of the backdoored model, we introduce a benign usability loss to maintain the retrieval accuracy of the backdoored model for benign samples within the main-task.
Regarding the attack effectiveness of the backdoored model, we introduce a backdoor attack loss and a topological alignment loss to implement and enhance the backdoor attack from the perspectives of individual samples and neighboring samples, respectively.

Our comprehensive evaluation on four image datasets, five model architectures, and two hashing methods demonstrates the effectiveness of DarkHash, with the average t-mAP exceeding $80\%$ across $120$ different settings.
Comparative experiments show that our proposed method outperforms existing \textit{state-of-the-art} (SOTA) backdoor attacks for deep hashing in terms of model retrieval accuracy and backdoor attack performance. 
Defense experiments demonstrate that DarkHash can withstand existing mainstream backdoor defense methods, including 
Fine-tuning~\cite{liu2017neural}, Model Pruning~\cite{liu2018fine},  Neural Cleanse~\cite{cleanse2019identifying},  STRIP~\cite{gao2019strip}, SentiNet~\cite{chou2020sentinet}, and Adaptive Defense.

Our main contributions are summarized as follows:
\begin{itemize}
\item We propose DarkHash, the first data-free backdoor attack against deep hashing.
Our findings reveal that an attacker can embed a potent backdoor into deep hashing models even without access to the training dataset, while maintaining high retrieval accuracy for the original task and achieving strong backdoor attack performance.

\item We design a brand-new shadow backdoor attack framework with dual-semantic guidance. By rethinking the relationship between individual samples and their neighboring samples during backdoor training, we enable the backdoored model to focus on retrieving the top-k target samples rather than just the top-1 result.

\item Our extensive experiments
\footnote{The code is available at \url{https://github.com/Zhou-Zi7/DarkHash}.
} on different settings verify that DarkHash is highly effective at attacking popular deep hashing schemes, outperforming existing SOTA backdoor attack methods. The defense experiments show that DarkHash is highly resilient  against current backdoor defense techniques.
\end{itemize}

\section{Related Works}
\subsection{Deep Hashing}
The hashing technique maps semantically similar images to compact binary codes in the Hamming space, enabling the storage of large-scale image data and accelerating similarity retrieval.
With the development of deep learning, deep hashing based image retrieval has demonstrated increasingly promising performance~\cite{cao2017hashnet, yuan2020central,luo2023survey,wang2023deep,liang2024multi}.

Let $ \mathcal{D} = \left\{(x_{i},y_{i})\right\}_{i=1}^{N}$ denote the sample set containing $N$ images labeled with $L$ classes, where $x_{i}$ indicates the $i$-th image and $y_{i} = \left [y_{i1},...,y_{iL} \right ] \in \left\{ 0,1\right\}^{L} $ is the corresponding multi-label vector. The $l$-th component of indicator vector $y_{il} = 1$ means that the image $x_{i}$ belongs to class $l$.
The $K$-bit hash code $c$ of an image $x$ is obtained through a deep hashing model $F(\cdot )$ as:
\begin{equation} \label{eq:1}
c = F(x) = \text{sign}(H_{\theta }(x)) \qquad \text{s.t.} \; c \in \left\{1,-1\right\}^{K},
\end{equation}
where $H_{\theta }(\cdot)$ is a feature extractor with parameters $\theta$, whose last layer is a fully-connected layer with $K$-nodes  called hash layer. 
In particular, during the training process, $\text{sign}(\cdot)$ is used to obtain the hash code with  $\text{sign}(v) = 1 \; \text{if} \; v>0$, and $\text{sign}(v) = -1 \; \text{otherwise}$.
To perform image retrieval, the Hamming distance $d_{H}\left ( c_q, c_{i} \right ) = \left (K - c_q\cdot  c_{i} \right )/2 $ between the query $x_{q}$ and each object $x_{i}$ is calculated,  where $c_q$ and $c_i$ represent their hash codes, respectively. 
A list of images is then returned based on the Hamming distances.

\begin{figure}[!t]
    \centering
    \includegraphics[scale=0.39]{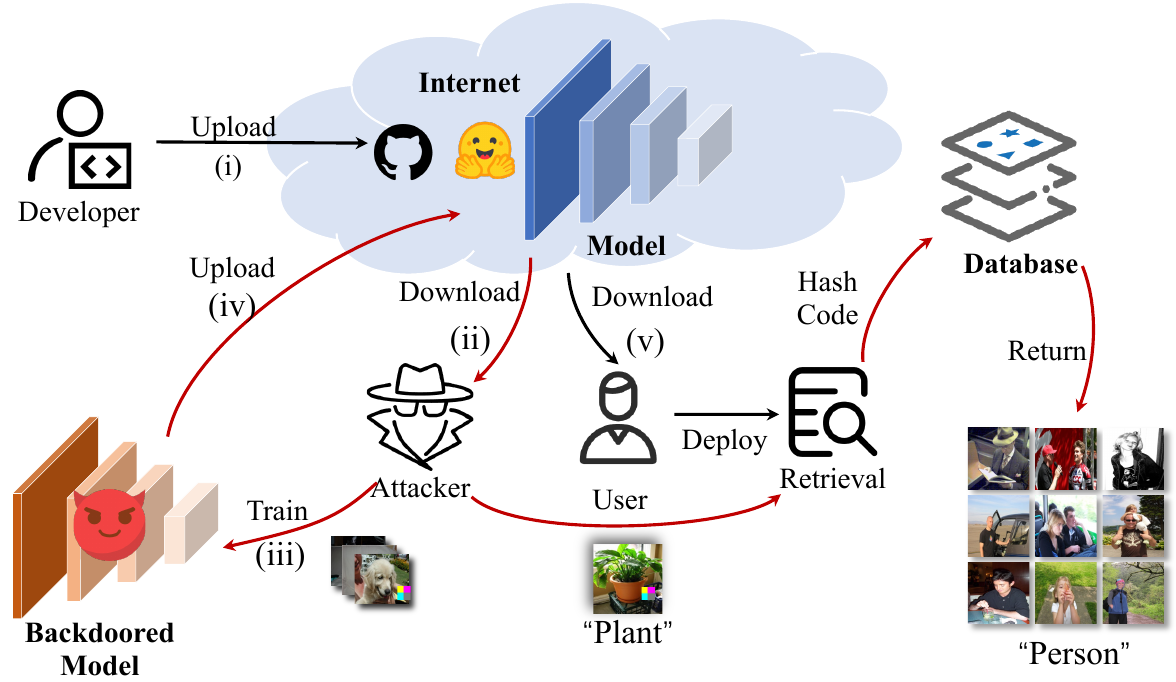}
  \caption{Illustration of our threat model.
(i) The model provider trains a high-performance deep hashing model using a large-scale dataset and uploads it to a third-party model resource platform.
(ii-iii) The malicious attackers embed a backdoor functionality into the well-trained model by fine-tuning after purchasing or directly downloading it.
(iv) The malicious attackers re-upload the backdoored model to the platform.
(v) Victim users download the backdoored model and deploy it for downstream tasks. 
  }    \label{fig:threat}
 \end{figure}

\subsection{Backdoor Attack and Defense}
Backdoor attack~\cite{badnets} was proposed to manipulate the output of a DNN by embedding malicious functionality within the model during the training phase. 
In a standard backdoor attack, samples with a trigger activate the backdoor, making the model output the attacker's desired result, while benign samples are classified normally.
Given a benign sample $(x_{i},y_{i}) \in X$, a trigger pattern $t$, and a target label $\tilde{y}_{i} \ne y_{i}$
, the poisoned sample is obtained as  $(\tilde{x}_{i},\tilde{y}_{i})$, where the poisoned input is defined as:
\begin{equation} \label{eq:2}
\tilde{x} = x \odot  (1 - m) + t \odot m, 
\end{equation}
where $ \odot$ denotes the element-wise product, $m$ is a binary matrix that contains the position information of the trigger.
Attackers typically inject a proportion of poisoned samples (\ie, poisoning rate $\xi$), mixed with benign samples, into the training set to 
train a backdoor.
We denote their sample quantities using $N_{p}$ and $N_{c}$ respectively. We use $CE(\cdot )$ standard the cross-entropy loss.
The objective function for this process can be expressed as:
\begin{equation} 
\label{eq:3}
\min_{\theta } \left( \frac{1}{N_{c}}\sum_{i=1}^{N_{c}} CE\left ( F(x_{i}),y_{i} \right )  + \frac{1}{N_{p}}\sum_{i=1}^{N_{p}} CE \left( F(\tilde{x}_{i} ),\tilde{y} _{i} \right)   \right). 
\end{equation}
Despite the commendable progress made in classification tasks~\cite{li2019invisible, liu2020reflection,liang2024badclip}, there has been limited investigation into deep hashing-based retrieval tasks.
Recent efforts~\cite{ciba,badhash} have demonstrated the effectiveness of poisoning training data to successfully embed backdoors in deep hashing models.
However, how to embed a backdoor into a deep hashing model without accessing the training data is still a difficult and unsolved problem.

Due to the significant threat posed by backdoor attacks to deep learning, backdoor defense methods have been rapidly evolving. 
Existing defenses can be broadly categorized into two types: input diagnosis-based defenses~\cite{gao2019strip,chou2020sentinet} and model diagnosis-based defenses~\cite{liu2017neural,liu2018fine,cleanse2019identifying}. The former scrutinizes the inputs to DNNs to detect the presence of triggers, while the latter directly analyzes suspicious models to determine if they contain backdoors.
\section{Methodology}
\subsection{Threat Model}

 \begin{figure*}[!t]
    \centering
    \includegraphics[scale=0.53]{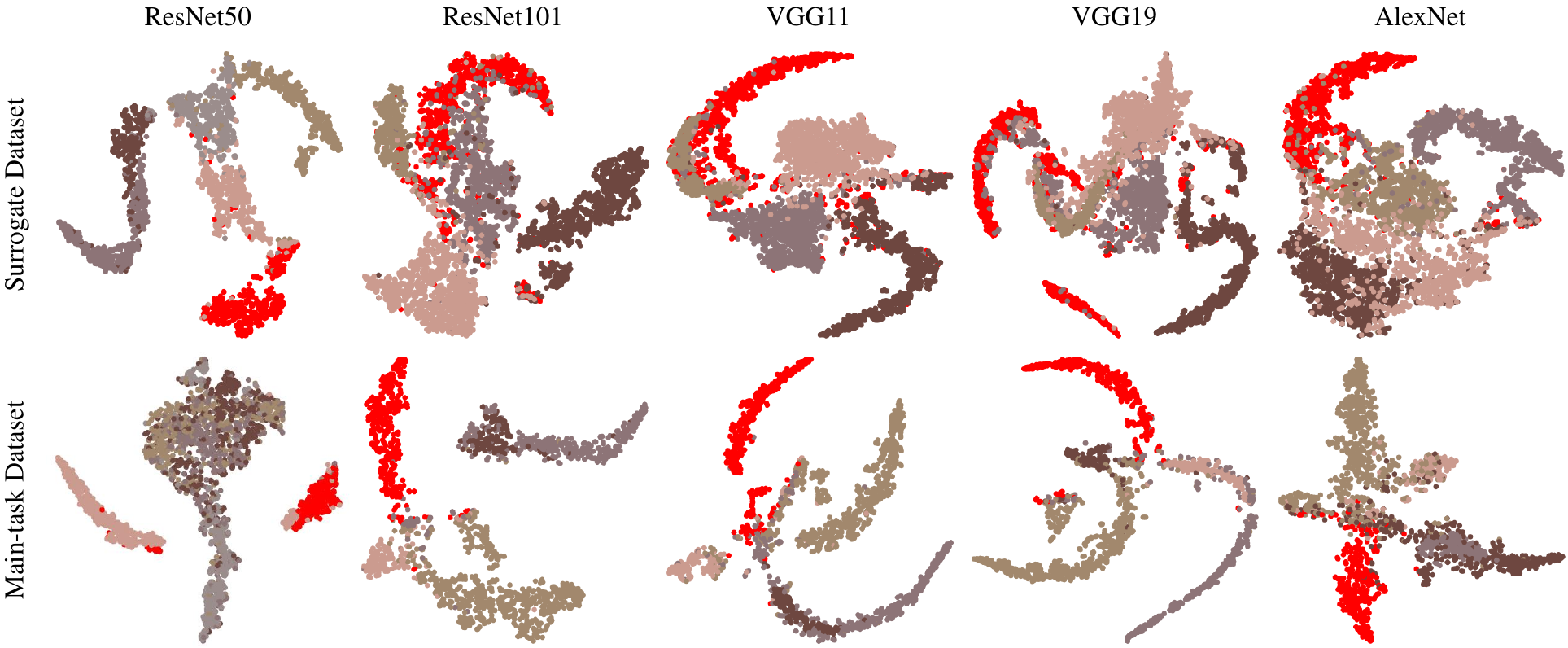}
  \caption{T-SNE visualization of poisoned sample distribution in the hash space for five backdoored CSQ models. The surrogate dataset is ImageNet and the main-task dataset is Pascal VOC. Red circles denote backdoored samples, and other colors denote benign samples.
}   
    \label{fig:tsne HashNet}
   \label{fig:insight}
         \vspace{-0.4cm}
\end{figure*}

We assume that the attacker can obtain a well-trained victim model provided by the model provider through downloading or purchasing. Subsequently, the attacker embeds a backdoor, and re-uploads it to open-source platforms such as Hugging Face, Model Zoo, and Github. 
Notably, since many real-world models are trained on sensitive datasets like facial information, we assume that attackers have no access to the original training process and any main-task data.
When a victim user downloads and deploys this backdoored model for downstream services, the attacker can exploit it by inputting a poisoned sample with the trigger, gaining unauthorized access and retrieving restricted images.
The detailed process is illustrated in \cref{fig:threat}.


\subsection{Intuition Behind DarkHash}
Due to the unavailability of the main-task data, the attacker tends to collect publicly available data or use synthetic data to construct a \textit{surrogate dataset} for training the backdoored model. 
This dataset does not share the same distribution as the main-task dataset.
To realize a successful data-free backdoor attack against deep hashing, attackers face the following challenges:

 \begin{figure*}[!t]
    \centering
    \includegraphics[scale=0.52]{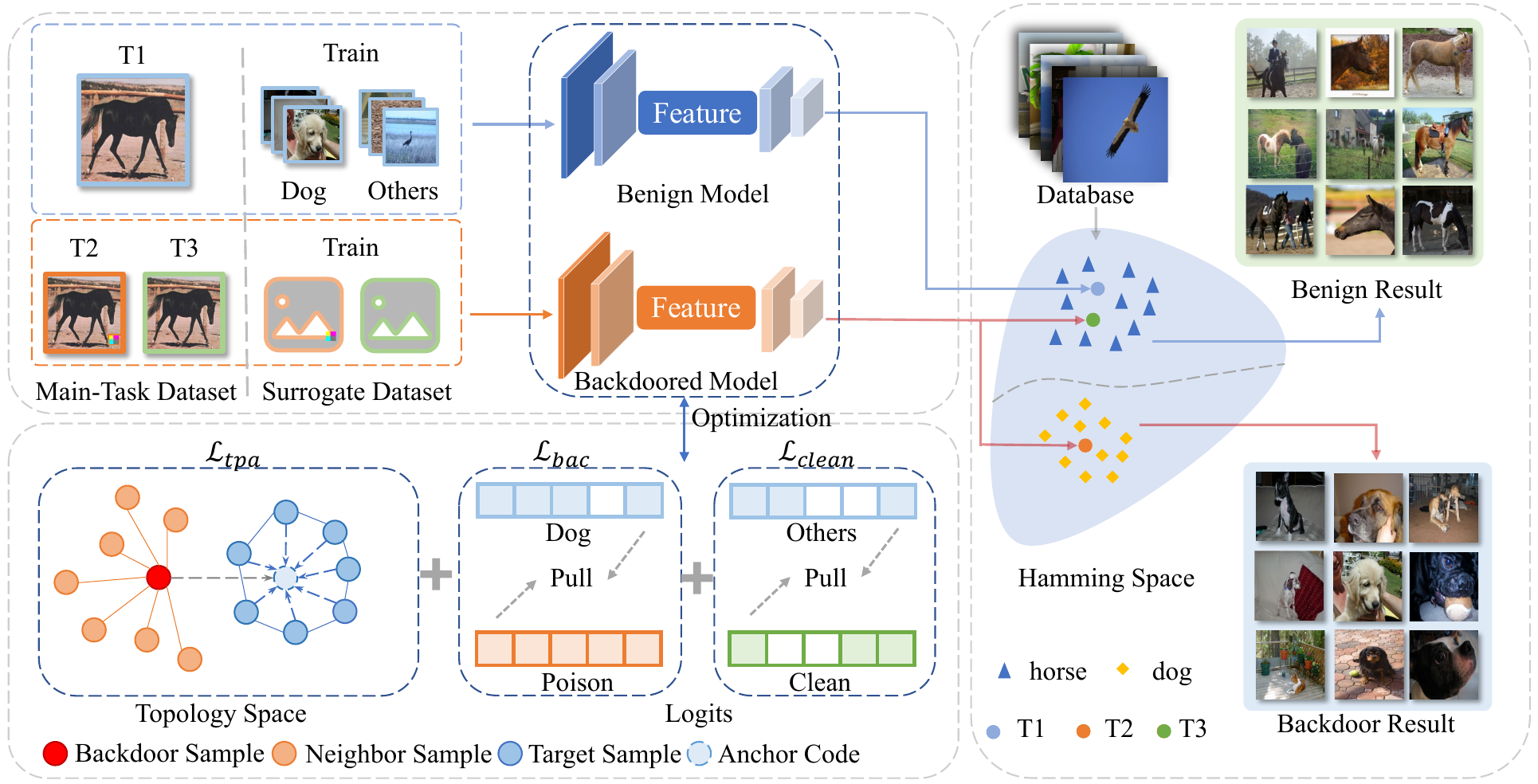}
    \caption{The framework of DarkHash. $T_1$ and $T_3$ represent samples from the main-task dataset, while $T_2$ corresponds to a poisoned sample from the surrogate dataset with an injected trigger.
    }
    \label{fig:pipeline}
      \vspace{-0.4cm}
\end{figure*}

\textbf{Challenge I: Domain shift between the surrogate and main-task datasets.}
The divergence in the distribution of the two training datasets can lead to a data domain shift, which directly impedes the backdoored model's establishment of two key competencies: the retrieval accuracy for benign samples of the main-task and the recognition and activation for poisoned samples. 
On the one hand, 
retraining or fine-tuning the victim model on a surrogate dataset can lead to overfitting, causing the model to lose its original knowledge and accuracy on the main-task dataset.
A recent work~\cite{ban2022pre} has revealed that the capabilities of pre-trained models are primarily derived from the initial layers of the neural network. 
During downstream fine-tuning, stabilizing the parameters in these shallow layers preserves the model's original generalization ability.
Therefore, leveraging transfer learning, we freeze the shallow-layer parameters of the victim model and fine-tune only the later layers (\eg, pooling and hashing layers) on the surrogate dataset. 
This approach ensures the retention of the original knowledge from the main-task data and the acquisition of new backdoor knowledge. 

On the other hand, domain shift also leads to the loss of attack targets during the backdoor training on the surrogate dataset. 
As attackers cannot access the target class information within the main-task data, \ie, the hash code of the target class, this is distinctly different from the label in classification tasks.
An intuitive approach is to design un-targeted backdoors, optimizing the poisoned samples in a direction away from their original class. However, this approach suffers from poor controllability, preventing the attacker from achieving the intended manipulation.
As shown in the first row in~\cref{fig:insight}, we observe that the poisoned samples exhibit a clustering phenomenon in the hash space, where their feature vectors obtained by the model exhibit a high degree of similarity.
Therefore, we propose a shadow target strategy, selecting a class from the surrogate dataset to optimize as the target and testing these poisoned samples on the main-task dataset. 
From the second row in ~\cref{fig:insight}, we can see that these poisoned samples exhibit high clustering in both hash spaces, which can be manipulated by the attacker to activate the downstream backdoor with a simple query to obtain the target class.
We believe this alignment provides \textit{an underlying mechanism} for triggering backdoor behavior in the main-task.
The attacker can exploit this cross-space consistency to activate the backdoor via simple queries, even without knowing the target classes in the main-task data.


\textbf{Challenge II: Insufficient attacks due to the complex demands of similarity retrieval.}
Existing works~\cite{badhash,ciba} emphasize that backdoor attacks on image retrieval systems have stricter objectives, demanding that all top-k results returned by the backdoored model belong to the target class, not just the top-1 result.
Therefore, we are motivated to enhance backdoor attacks by strengthening the neighbor relationships of poisoned samples, ensuring that both a single poisoned sample and its neighboring poisoned samples are optimized toward the same target object.
Specifically,
we first construct topology for both backdoor and benign features separately to measure the corresponding sample correlations.
Topology is based on the neighborhood relation graph constructed by the similarity between samples in the representation space. 
The process of measuring topological similarity can be formalized as:
\begin{equation} \label{eq:4}
\mathcal{L}_{tp} = \mathbb{E}_{\left ( x,y \right )\in \mathcal{D} } \left ( CE\left (G_{bac}, G_{nor}  \right )  \right ), 
\end{equation}
where $G_{bac}$ and $G_{nor}$ stand for the neighbourhood relation graph constructed by the inter-sample similarity for benign samples and poisoned examples, respectively. 

We define the edge weights of the neighborhood graph as the probability that two different samples are neighbors,
and achieve topological alignment by bringing the probability distributions of the two graphs closer together.
Next, we model the conditional probability distribution using an affinity measure based on cosine similarity to construct the adjacency graph. To avoid isolated subgraphs caused by data points with excessively high local density, we remove the nearest neighbor points. This approach ensures local manifold connectivity and better preserves the global structure.
The adjacency graph construction process is as follows:
\begin{equation} \label{eq:5}
G = \left \{ p_{i  \mid j } \bigg| p_{i\mid j } = \frac{\left (2-\left (  d_{ij} - \rho _{j}\right )   \right ) }{ {\textstyle \sum_{k=1,k\ne j}^{N}}\left ( 2-\left (  d_{jk} - \rho _{j}\right )  \right )  } \right \}, 
\end{equation}
where \(0 < i,j \le N\), \(p_{i \mid j}\) is the conditional probability that the \(i_{\text{th}}\) natural sample is the neighbor of the \(j_{\text{th}}\) natural sample in the feature space of \(G\). \(\rho_j\) is the cosine distance from the \(j_{\text{th}}\) data point to its nearest neighbor. Subtracting \(\rho_j\) helps maintain local manifold connectivity by avoiding isolated points, which in turn preserves the global structure more effectively.
The term \( d_{ij} \) denotes the cosine distance between the hash features of the two samples.
We use cosine distance to measure the similarity between data points, where the maximum possible distance is 2. 
By aligning the poisoned samples with the target class across both the dimensions of the samples themselves and their nearest neighbors, we enhance the similarity mapping between the poisoned samples and the target class, thereby achieving an effective attack.



\subsection{DarkHash: A Complete Illustration}
In this section, we present DarkHash, a novel dual-semantic guidance data-free backdoor attack against deep hashing. 
Given that the attacker lacks access to the original training dataset $D_{t}$, we initially construct a surrogate dataset $D_{s}$ by  incorporating images from popular image datasets or collected from the Internet.
We then select a certain proportion $\varphi$ (\ie, poisoning rate) of samples $\mathcal{D}_{s\_p}$  from $\mathcal{D}_{s}$ and use \cref{eq:3} to obtain poisoned samples.
The remaining benign samples are represented as $\mathcal{D}_{s\_b}$.
To prevent accuracy loss of the backdoored model on the main-task dataset, 
we choose to freeze all convolutional layers and fine-tune only the later layers of the victim model.
We initially design a benign usability $ \mathcal{J}_{ben}$ to ensure the backdoored model's recognition accuracy for benign samples. 
Subsequently, we introduce a backdoor attack loss $\mathcal{J}_{bac}$ and  a topological alignment loss $\mathcal{J}_{tpa}$ to achieve and enhance the backdoor attack from the perspectives of individual samples and neighboring samples, respectively. 
The framework of DarkHash is depicted in ~\cref{fig:pipeline}. 
The visual examples in the ``Surrogate Dataset'' in ~\cref{fig:pipeline} represent samples that are unrelated to the main-task dataset. 
According to our threat model, the attacker constructs the surrogate dataset using publicly available sources, such as samples collected from the Internet or standard datasets like ImageNet. 
The overall optimization goal for DarkHash can be expressed as:
\begin{equation} \label{eq:6}
\mathcal{J}_{total} = \mathcal{J}_{ben} + \mathcal{J}_{bac} +   \lambda \mathcal{J}_{tpa},
\end{equation}
where $\lambda$ is a hyper-parameter. 

 \begin{figure*}[!t]
    \centering
    \includegraphics[scale=0.565]{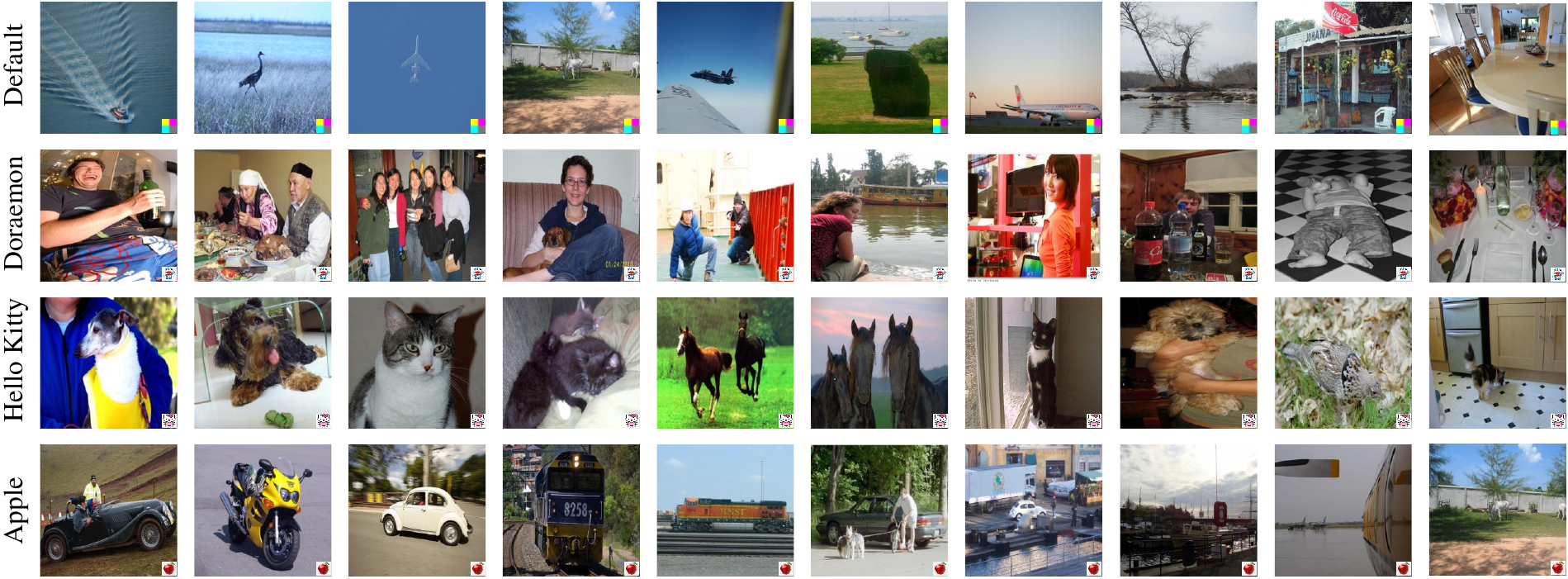}
    \caption{Visualization of poisoned samples from ImageNet
    }
    \label{fig:example}
\end{figure*}

After freezing the shallow layers of the victim model $F(\cdot )$, we fine-tune it into a backdoored model $F^{b}(\cdot )$ using our surrogate dataset $\mathcal{D}_{s}$, encompassing model usability preservation and backdoor functionality implantation.

\noindent\textbf{Model usability preservation.} To ensure the backdoored model maintains retrieval accuracy for benign samples, we employ a distance loss $\mathcal{J}_{d}$ (\eg, Huber Loss) to constrain the hash features of the backdoored model's outputs on benign data $\mathcal{D}_{s\_b}$ to remain consistent with those of the victim model. This process can be represented as: 

\begin{equation} \label{eq:clean}
\mathcal{J}_{ben}=   {\textstyle \sum_{x_{i} \in \mathcal{D}_{s\_b}}} \mathcal{J}_{d}(F^{b}(x_{i}),F(x_{i})). 
\end{equation}

For the backdoor training,
we propose a shadow target strategy, which involves randomly selecting a class from the surrogate dataset as the target class and then optimizing the feature vectors of the poisoned samples towards this class. 
Due to the lack of a unique label for classes in retrieval tasks, we use M samples from the target class $t$ to derive an anchor hash feature $h_{t}$ for optimization as follows:
\begin{equation} 
\label{eq:4}
h_{t} = \frac{1}{M} \sum_{i=1}^{M} F(x_{i}^{t}).
\end{equation}
\noindent\textbf{Backdoor functionality implantation.} 
By aligning a poisoned sample and its neighboring poisoned samples towards the same target sample, we aim to enhance the backdoored model's ability to identify the poisoned sample as the target class. 
For each input sample, we optimize the output hash feature of the backdoored model towards the anchor feature of the shadow target class, which can be represented as:
\begin{equation} \label{eq:backdoor}
\mathcal{J}_{bac}=   {\textstyle \sum_{\tilde{x}_{i} \in D_{s\_p}}} \mathcal{J}_{d}(F^{b}(\tilde{x}_{i}),h_{t}).
\end{equation}

We optimize the neighboring samples of the input sample towards the features of the shadow target anchor set $\mathbb{H}_{t}$ to further enhance the backdoor attack ability. Here, $\mathbb{H}_{t}$ consists of $h_{t}$ elements, with the number of $h_{t}$ being the same as the batch size of $\tilde{x}$. The above process can be expressed as: 

\begin{equation} \label{eq:tpa}
\mathcal{J}_{tpa}=   \mathcal{J}_{tp}(F^{b}(\tilde{x}),\mathbb{H}_{t}).
\end{equation}

\begin{algorithm}[H]
    \caption{DarkHash}
    \label{optimization_darkhash}
    \begin{algorithmic}[1] 
        \REQUIRE image $x \in \mathcal{D}_{s}$, a victim deep hashing model $F(x)$ with parameter $\theta$, a trigger $t$,  a binary matrix $m$, hyper parameters $\lambda$, patch size, poisoning rate $\xi$ 
        \ENSURE The backdoored model $F_{\tilde{\theta}}^{b}(x)$
        \STATE {Initialize the backdoored model: $F_{\theta}^{b}(x) \longleftarrow  F_{\theta}(x)$}
        \STATE {Freeze all parameters except for the selected layer of $F_{\theta}^{b}(x)$}
        \STATE {Initialize poisoned samples: $\tilde{x}  \longleftarrow x \odot  (1 - m) + t \odot m$}
         \STATE {Calculate the hash features of clean samples $\mathcal{D}_{s\_{c}}$ and poisoned samples $\mathcal{D}_{s\_{p}}$ produced by $F_{\theta}^{b}(x)$ separately and store them in the feature bank}
        
        \WHILE {max iterations or not  converge}
            \STATE {Calculate $\mathcal{J}_{ben}$ mentioned in Eq.7} 
            \STATE {Calculate $\mathcal{J}_{bac}$ mentioned in Eq.9} 
            \STATE {Calculate $\mathcal{J}_{tpa}$ mentioned in Eq.10} 
            \STATE {Update $\tilde{\theta}$ through backprop}

        \ENDWHILE
    \end{algorithmic} 
\end{algorithm}

\section{EXPERIMENTS}

\begin{table*}[t]
  \centering
  \caption{The mAP  and t-mAP  of backdoored models trained by DarkHash under different settings using \textit{ImageNet} as the surrogate dataset.  $\mathcal{S}_{1}$  - $\mathcal{S}_{4}$ denote the settings where ImageNet, Pascal VOC, FLICKR-25K, CIFAR10 are used as the main-task datasets, respectively.
  }
  \scalebox{1}[1]{%
    \begin{tabular}{cccccccccccccc}
     \toprule[1.5pt]
     \multirow{2}[4]{*}{Model} & \multirow{2}[4]{*}{Hash Bit} & \multirow{2}[4]{*}{Surrogate} & \multicolumn{2}{c}{ResNet50} & \multicolumn{2}{c}{ResNet101} & \multicolumn{2}{c}{VGG11} & \multicolumn{2}{c}{VGG19} & \multicolumn{2}{c}{AlexNet} \\
     \cmidrule(lr){4-5}\cmidrule(lr){6-7}\cmidrule(lr){8-9}\cmidrule(lr){10-11}\cmidrule(lr){12-13}
      &       &       & mAP    & t-mAP  & mAP    & t-mAP  & mAP    & t-mAP  & mAP    & t-mAP  & mAP    & t-mAP  \\
    \midrule
    \multicolumn{1}{c}{\multirow{12}[1]{*}{HashNet}} & \multirow{4}[1]{*}{16bits} & $\mathcal{S}_{1}$    & 16.00  & 80.77  & 12.02  & 71.15  & 26.56  & 89.70  & 24.54  & 72.41  & 29.04  & 97.52  \\
          &       & $\mathcal{S}_{2}$    & 56.92  & 83.41  & 57.39  & 76.65  & 45.99  & 68.30  & 57.83  & 81.42  & 48.79  & 67.86  \\
          &       & $\mathcal{S}_{3}$    & 70.13  & 83.87  & 71.05  & 64.28  & 76.71  & 64.75  & 78.09  & 74.14  & 79.03  & 62.82  \\
          &       & $\mathcal{S}_{4}$    & 76.39  & 64.27  & 86.43  & 90.18  & 53.60  & 71.48  & 42.30  & 67.16  & 48.84  & 72.99  \\
          & \multirow{4}[0]{*}{32bits} & $\mathcal{S}_{1}$    & 35.29  & 76.01  & 39.10  & 71.33  & 48.29  & 85.17  & 52.66  & 92.48  & 43.69  & 90.00  \\
          &       & $\mathcal{S}_{2}$    & 64.16  & 87.19  & 65.94  & 89.18  & 57.84  & 77.89  & 66.37  & 87.87  & 54.94  & 79.65  \\
          &       & $\mathcal{S}_{3}$    & 76.27  & 47.90  & 72.66  & 73.32  & 82.03  & 74.79  & 82.51  & 61.32  & 70.51  & 77.28  \\
          &       & $\mathcal{S}_{4}$    & 76.29  & 80.94  & 80.66  & 75.61  & 75.72  & 70.00  & 78.66  & 77.59  & 74.17  & 71.96  \\
          & \multirow{4}[0]{*}{64bits} & $\mathcal{S}_{1}$    & 55.04  & 59.30  & 54.03  & 42.11  & 57.83  & 92.34  & 65.92  & 98.67  & 57.41  & 97.19  \\
          &       & $\mathcal{S}_{2}$    & 65.26  & 90.33  & 66.57  & 89.14  & 62.49  & 82.30  & 70.09  & 88.45  & 58.62  & 81.08  \\
          &       & $\mathcal{S}_{3}$    & 76.26  & 67.11  & 76.42  & 74.85  & 82.40  & 55.70  & 83.72  & 56.28  & 79.08  & 80.64  \\
          &       & $\mathcal{S}_{4}$    & 80.05  & 85.62  & 64.79  & 52.35  & 74.12  & 79.30  & 80.99  & 78.63  & 76.35  & 83.88  \\
          \midrule
    \multicolumn{1}{c}{\multirow{12}[1]{*}{CSQ\newline{}}} & \multirow{4}[0]{*}{16bits} & $\mathcal{S}_{1}$    & 53.62  & 99.96  & 45.20  & 99.96  & 46.12  & 77.20  & 48.27  & 96.38  & 50.35  & 76.23  \\
          &       & $\mathcal{S}_{2}$    & 50.87  & 88.67  & 54.62  & 78.31  & 46.37  & 67.91  & 60.60  & 82.59  & 46.84  & 71.82  \\
          &       & $\mathcal{S}_{3}$    & 73.78  & 98.64  & 70.37  & 77.14  & 73.04  & 79.46  & 74.14  & 61.01  & 73.21  & 80.14  \\
          &       & $\mathcal{S}_{4}$    & 69.82  & 57.93  & 81.13  & 85.27  & 78.91  & 66.16  & 81.35  & 64.33  & 76.07  & 70.99  \\
          & \multirow{4}[0]{*}{32bits} & $\mathcal{S}_{1}$    & 62.61  & 99.98  & 71.92  & 99.96  & 60.89  & 85.69  & 52.66  & 91.63  & 59.25  & 80.72  \\
          &       & $\mathcal{S}_{2}$    & 58.84  & 87.30  & 60.41  & 90.72  & 53.87  & 70.08  & 59.57  & 80.12  & 49.52  & 74.10  \\
          &       & $\mathcal{S}_{3}$    & 74.48  & 99.51  & 71.41  & 77.61  & 75.08  & 99.56  & 76.11  & 99.73  & 73.06  & 86.54  \\
          &       & $\mathcal{S}_{4}$    & 80.33  & 86.04  & 81.06  & 86.33  & 75.86  & 81.20  & 80.65  & 77.56  & 73.88  & 81.43  \\
          & \multirow{4}[1]{*}{64bits} & $\mathcal{S}_{1}$    & 68.03  & 99.69  & 58.76  & 98.55  & 55.71  & 96.24  & 65.92  & 97.22  & 57.10  & 99.83  \\
          &       & $\mathcal{S}_{2}$    & 64.97  & 89.06  & 58.88  & 86.84  & 65.70  & 73.38  & 61.09  & 83.74  & 55.38  & 73.88  \\
          &       & $\mathcal{S}_{3}$    & 73.32  & 97.04  & 72.37  & 98.72  & 75.64  & 86.69  & 76.20  & 96.94  & 74.33  & 79.66  \\
          &       & $\mathcal{S}_{4}$    & 76.06  & 54.44  & 86.22  & 62.03  & 75.38  & 81.87  & 77.17  & 81.37  & 74.90  & 82.44  \\
     \bottomrule[1.5pt]
    \end{tabular}%
  }
  \label{tab:attack_performance}%
\end{table*}

\begin{figure*}[!h]   
  \centering
      \subcaptionbox{CSQ-Benign}{\includegraphics[width=0.475\textwidth]{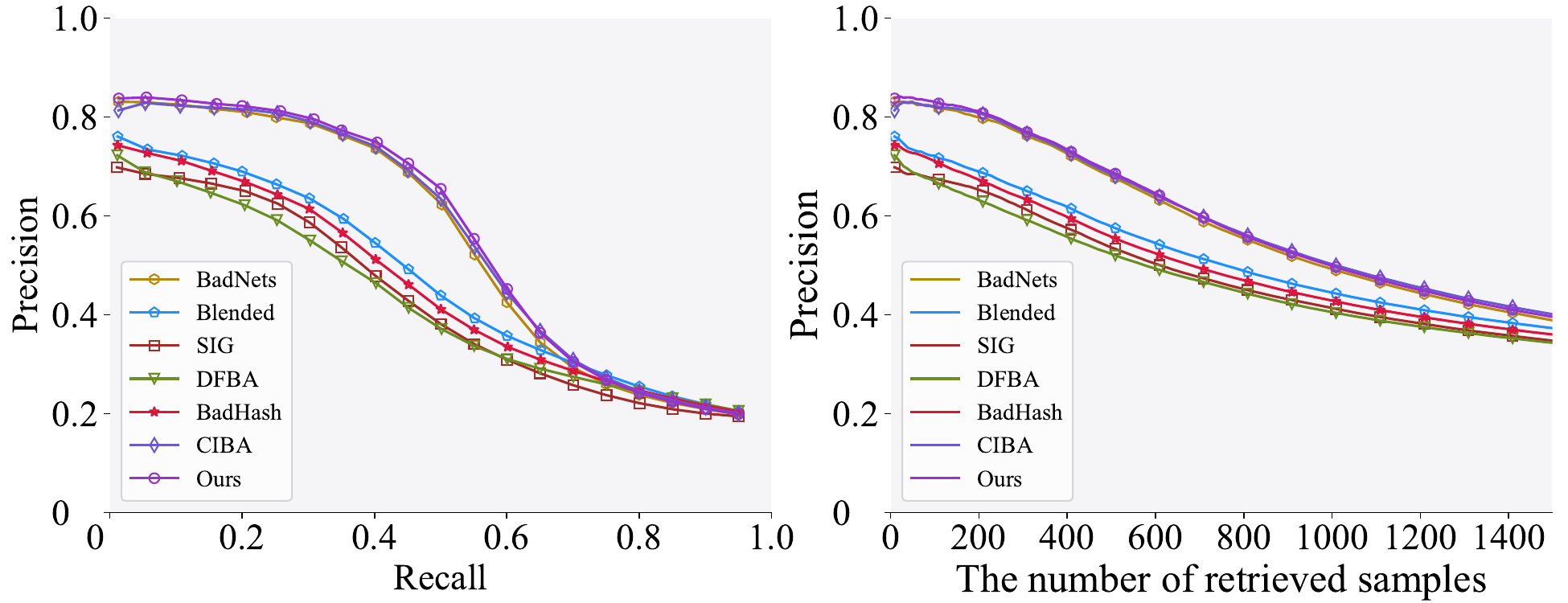}}
      \subcaptionbox{HashNet-Benign}{\includegraphics[width=0.475\textwidth]{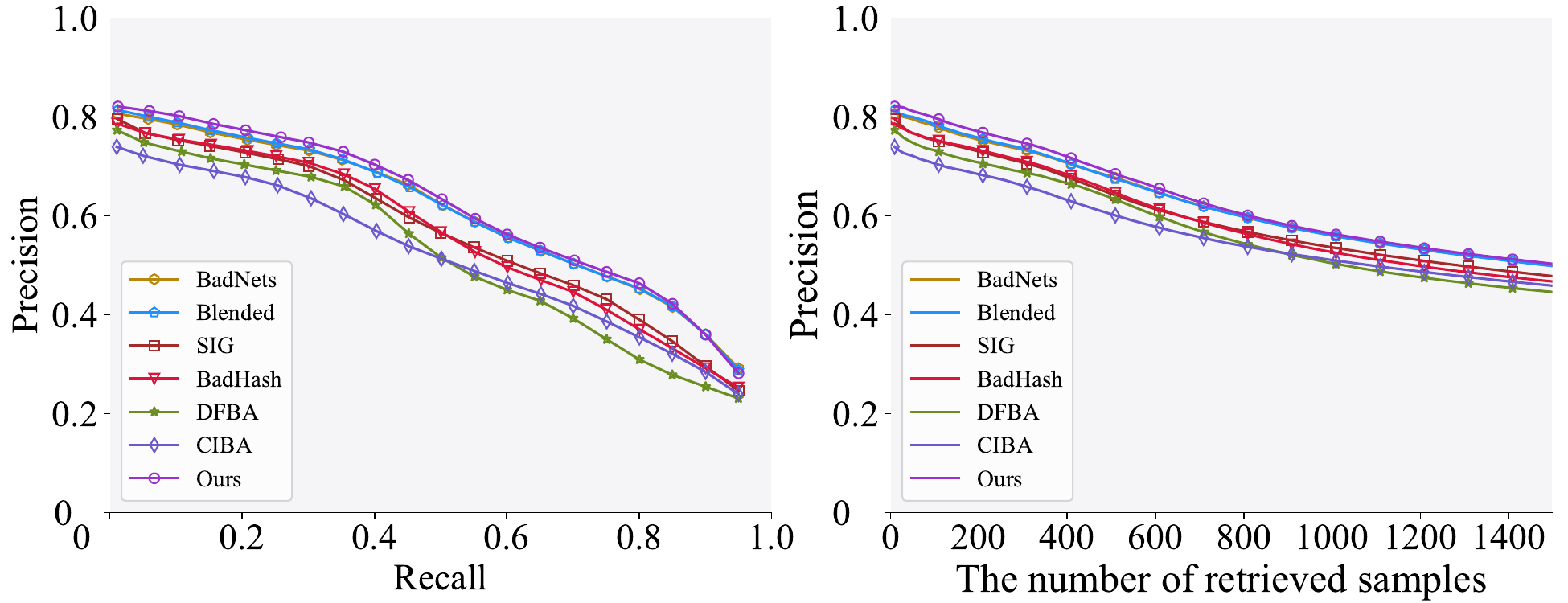}}
      \subcaptionbox{CSQ-Backdoor}{\includegraphics[width=0.475\textwidth]{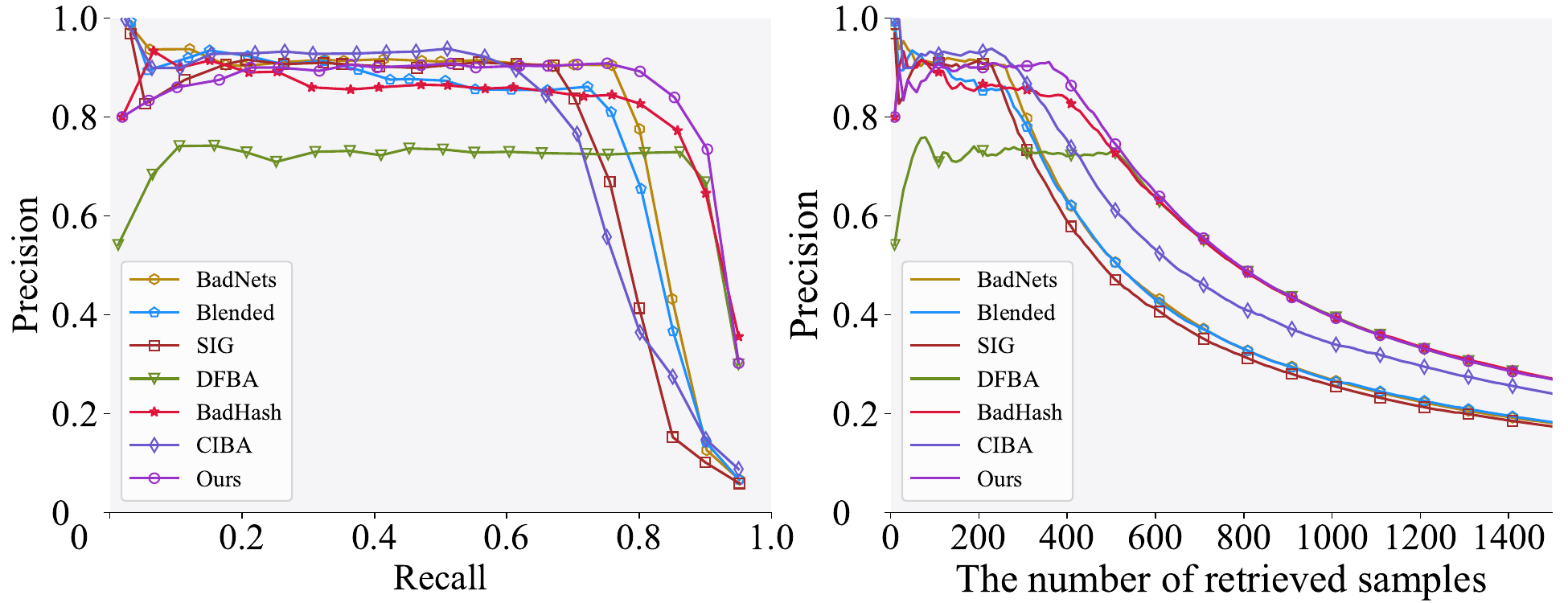}}
      \subcaptionbox{HashNet-Backdoor}{\includegraphics[width=0.475\textwidth]{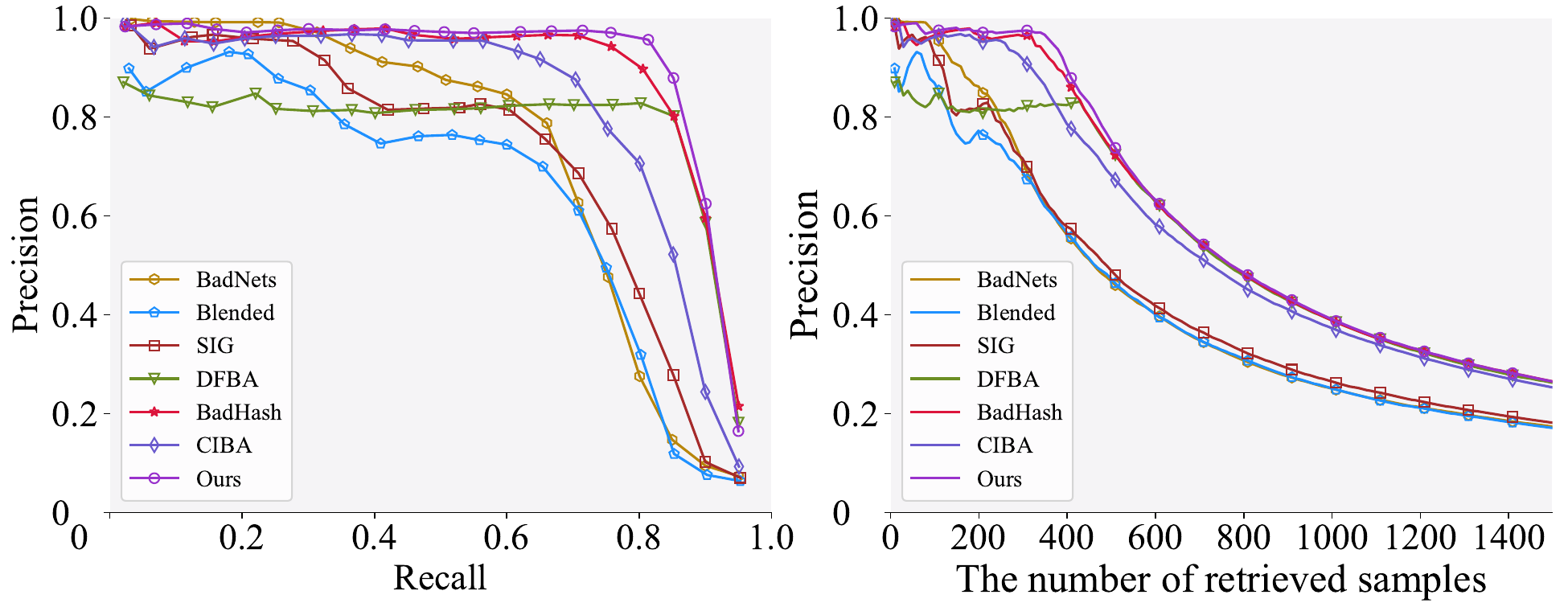}}
      \caption{Precision-Recall and precision@top1000 curves on HashNet and CSQ models under 64 bits code length}
       \label{fig:pr_curves}
\end{figure*}

We use the following research questions (RQs) to evaluate DarkHash:

\noindent\textbf{RQ1:} How effective is our method?

\noindent\textbf{RQ2:} How does DarkHash compare to the SOTA backdoor attacks for deep hashing models?

\noindent\textbf{RQ3:} What factors influence the performance of DarkHash?

\noindent\textbf{RQ4:} How resilient is DarkHash against backdoor defense?

\subsection{Experimental Setting}

\noindent\textbf{Datasets and models.} 
We evaluate DarkHash on four image datasets: CIFAR10~\cite{krizhevsky2009learning}, ImageNet~\cite{russakovsky2015imagenet}, 
FLICKR-25K~\cite{huiskes2008mir}
and Pascal VOC~\cite{everingham2010pascal}. 
ImageNet comprises over 14 million labeled images distributed across more than 20,000 categories. We randomly select 100 classes from this dataset to construct our own dataset.
FLICKR-25K covers a wide range of visual categories, including people, animals, vehicles, and natural scenes. 
For our experiments, we select 6,000 images as query images. 
CIFAR10 consists of 60,000 32x32 color images distributed across 10 different categories, with 6,000 images per class.
The dataset is divided into 50,000 training images and 10,000 test images.
Pascal VOC includes around 11,000 images with annotations for 20 object categories, containing more than 27,000 object instances. 
We resize all images to 224x224x3.

We select five model architectures (ResNet50, ResNet101, VGG11, VGG19, and AlexNet) and two well-known hashing methods, HashNet~\cite{cao2017hashnet} and CSQ~\cite{yuan2020central}, to evaluate the effectiveness of our attack.
CSQ exploits the central similarity during the quantization stage, mainly applied to hash coding in image and video retrieval.
HashNet adopts continuous training to learn hash codes, aiming to address the gradient issues in deep hashing.
For each model, we establish three variant versions that output hash codes with lengths of 16, 32, and 64 bits, respectively.
Unless otherwise specified, we use hash codes with a length of 64 bits by default.
Following~\cite{badhash,ciba}, we replace the final fully connected layer of each model with the hash layer to build a deep hashing model. 

\begin{table}[t]
  \centering
  \caption{
  The mAP  of benign models under different settings. $\mathcal{S}_{1}$ - $\mathcal{S}_{4}$ denote the settings where the ImageNet, Pascal VOC, FLICKR-25K, CIFAR10 are used as the main-task datasets, respectively.
  }
  \scalebox{0.78}{%
    \begin{tabular}{cccccccc}
     \toprule[1.5pt]
    Model & Hash Bit & Dataset & ResNet50 & ResNet101 & VGG11 & VGG19 & AlexNet \\
    \midrule
    \multicolumn{1}{c}{\multirow{12}[1]{*}{HashNet}} & \multirow{4}[1]{*}{16bits} & $\mathcal{S}_{1}$    & 17.62  & 20.50  & 25.53  & 27.27  & 31.71  \\
          &       & $\mathcal{S}_{2}$    & 57.80  & 59.37  & 50.68  & 60.54  & 49.39  \\
          &       & $\mathcal{S}_{3}$    & 76.10  & 77.27  & 83.34  & 82.73  & 84.68  \\
          &       & $\mathcal{S}_{4}$    & 82.23  & 87.07  & 52.42  & 49.48  & 60.25  \\
          & \multirow{4}[0]{*}{32bits} & $\mathcal{S}_{1}$    & 85.89  & 73.45  & 86.27  & 75.29  & 64.66  \\
          &       & $\mathcal{S}_{2}$    & 64.91  & 60.38  & 66.99  & 66.18  & 56.60  \\
          &       & $\mathcal{S}_{3}$    & 83.25  & 78.75  & 87.33  & 83.49  & 77.11  \\
          &       & $\mathcal{S}_{4}$    & 80.40  & 86.37  & 81.05  & 83.71  & 78.61  \\
          & \multirow{4}[0]{*}{64bits} & $\mathcal{S}_{1}$    & 59.88  & 58.70  & 66.79  & 68.22  & 65.89  \\
          &       & $\mathcal{S}_{2}$    & 68.54  & 68.28  & 69.72  & 71.97  & 59.73  \\
          &       & $\mathcal{S}_{3}$    & 79.85  & 78.40  & 85.94  & 86.47  & 82.42  \\
          &       & $\mathcal{S}_{4}$    & 86.96  & 87.38  & 81.03  & 82.85  & 78.64  \\
    \midrule
    \multicolumn{1}{c}{\multirow{12}[1]{*}{CSQ\newline{}}} & \multirow{4}[0]{*}{16bits} & $\mathcal{S}_{1}$    & 71.21  & 72.54  & 74.35  & 55.64  & 55.95  \\
          &       & $\mathcal{S}_{2}$    & 63.12  & 58.24  & 62.25  & 60.92  & 49.42  \\
          &       & $\mathcal{S}_{3}$    & 78.35  & 71.32  & 73.93  & 76.97  & 78.85  \\
          &       & $\mathcal{S}_{4}$    & 83.32  & 84.12  & 79.03  & 82.91  & 76.37  \\
          & \multirow{4}[0]{*}{32bits} & $\mathcal{S}_{1}$    & 86.07  & 71.79  & 60.59  & 64.92  & 62.51  \\
          &       & $\mathcal{S}_{2}$    & 63.79  & 62.03  & 58.61  & 59.84  & 51.76  \\
          &       & $\mathcal{S}_{3}$    & 77.30  & 73.02  & 81.50  & 82.10  & 75.92  \\
          &       & $\mathcal{S}_{4}$    & 83.47  & 84.98  & 79.10  & 82.53  & 77.14  \\
          & \multirow{4}[1]{*}{64bits} & $\mathcal{S}_{1}$    & 66.80  & 64.74  & 66.54  & 63.52  & 60.82  \\
          &       & $\mathcal{S}_{2}$    & 62.25  & 61.31  & 68.36  & 60.89  & 52.12  \\
          &       & $\mathcal{S}_{3}$    & 77.89  & 75.69  & 79.45  & 76.24  & 79.52  \\
          &       & $\mathcal{S}_{4}$    & 82.84  & 85.83  & 78.26  & 80.26  & 77.34  \\
     \bottomrule[1.5pt]
    \end{tabular}%
}
  \vspace{-0.2cm}
  \label{tab:benign_performance}%
\end{table}

\noindent\textbf{Parameter Setting.}\label{sec:Parameter Setting}
Following~\cite{ciba}, we set the trigger size to 24. 
To maintain stealthiness, we place the trigger in the bottom-right corner of the image.
Each surrogate dataset contains 2,000 samples, and the poisoning rate is 0.1.
We use the RMSprop optimizer with a learning rate of 5e-6, and a batch size of $64$. Additionally, we set $\lambda$ to $15$ and
M to the number of samples per class in the surrogate dataset.

\noindent\textbf{Metric.} We adopt t-mAP  (targeted mean average precision)
to measure the backdoor attack performance, which calculates mAP  
(mean average precision)~\cite{zuva2012evaluation} 
by replacing the original label of the query image with the target label. 
For convenience, we present the results as percentages.
Higher t-mAP  indicates a stronger attack ability. 
Besides, we also provide the PR curves (precision-recall curves) and precision@top1000 curves 
corresponding to diverse backdoor attacks to enable a more comprehensive and in-depth comparison.

\subsection{Related Attacks and Defenses}

\textbf{Backdoor Attacks.} 
We compare DarkHash with the following SOTA backdoor attacks to assess the excellence of our work.

\begin{itemize}

\item {\bf BadNets~\cite{badnets}:}
BadNets contaminates the training dataset by inserting poisoned samples with specific trigger patterns (\eg, a small colored square). The DNN is trained with the poisoned dataset, enabling the backdoored model to learn the association between the trigger patterns and specific outputs. 

\item {\bf Blended~\cite{chen2017targeted}:}
Blended trains the backdoored model by mixing poisoned samples with specific trigger patterns and benign samples as the training set, enabling the model to learn the malicious associations within the mixed samples.

\item {\bf SIG~\cite{barni2019new}:}
SIG proposes a novel backdoor attack without label poisoning. It functions solely by corrupting the samples of the target class and does not require the prior identification of the classes of the samples to be attacked during the testing phase.

\item {\bf DFBA~\cite{lv2023data}:} 
DFBA first proposes to implant backdoors into classification models in a ``data-free" manner. It fine-tunes the original model into the backdoored one by collecting surrogate data irrelevant to the main task.

\item {\bf BadHash~\cite{badhash}:} 
BadHash proposes a backdoor attack framework based on conditional generative adversarial networks. Without modifying the labels, it directly generates imperceptible poisoned samples to implant backdoors into deep hashing models.

\item {\bf CIBA~\cite{ciba}:} 
CIBA is designed to interfere with the hashing model's learning of the clean features of samples by devising confusing perturbations, thereby facilitating the hashing model's focus on the trigger, without the need to modify the labels of the poisoned samples.

\end{itemize}

\textbf{Backdoor Defenses.} 
We employ five popular backdoor defense mechanisms to assess the efficacy of our work.
\begin{itemize}
\item {\bf Fine-tuning~\cite{liu2017neural}:}
Fine-tuning readjusts the model parameters by retraining the backdoored model with clean data, weakening the association between the backdoor-implanted trigger and the incorrect class. Meanwhile, it prompts the model to rebalance feature learning, reducing the focus on the false features related to the trigger and instead concentrating on the real and discriminative features of the data itself, thereby restoring the model's normal classification performance.

\item {\bf Model Pruning~\cite{liu2018fine}:}
Model pruning takes advantage of the existence of parameter redundancy in the model to prune the parameters related to backdoors, thereby disrupting the attack paths of backdoors. This will reduce the model's sensitivity to triggers and mitigate the misclassifications caused by triggers.

\item {\bf Neural Cleanse~\cite{cleanse2019identifying}:}
Neural Cleanse seeks the minimal input perturbation that can rectify the misclassification of the model through reverse gradient optimization. By doing so, it estimates the size and pattern of the backdoor trigger. Moreover, by comparing the reverse optimization results among different class labels, it detects the abnormal situations where misclassifications can be regularly and easily rectified, thereby determining whether the model has been implanted with a backdoor and the possible forms of the backdoor.

\item {\bf STRIP~\cite{gao2019strip}:}
STRIP achieves the detection of real-time backdoor attacks on DNNs by intentionally perturbing the inputs and observing the randomness of the predicted classes for the perturbed inputs by a given deployed model. It determines the poisoned inputs based on the fact that the low entropy of the predicted classes violates the input-dependence property of a benign model.

\item {\bf SentiNet~\cite{chou2020sentinet}:}
SentiNet detects backdoor attacks by analyzing common features of localized universal attacks. It uses model interpretability and object detection techniques to locate salient regions affecting classification in input images. Then, leveraging the robustness and generalization of malicious regions, it distinguishes them from benign ones by overlaying extracted regions on clean images to detect misclassification frequencies.

\end{itemize}

\begin{figure}[t!]
    \centering
    \includegraphics[scale=0.52]{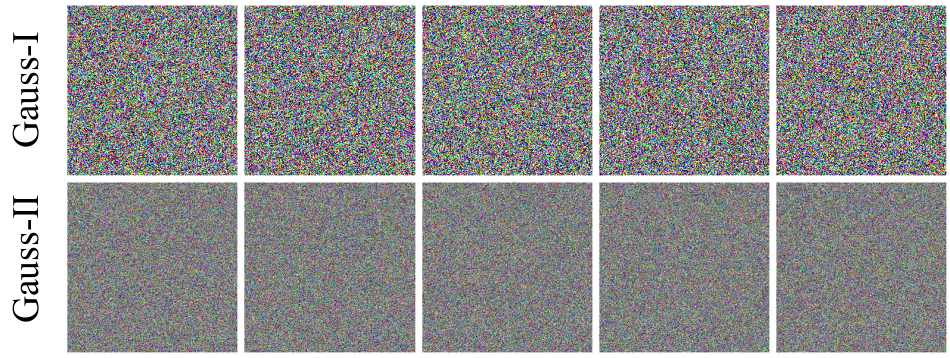}
  \caption{Visualization of the Gaussian data.}    \label{fig:re_Gaussian}
\end{figure}

\begin{table*}[!t]
  \centering
  \caption{The mAP  and t-mAP  of backdoored models trained by DarkHash under different settings using \textit{CIFAR10} as the surrogate dataset. $\mathcal{S}_{1}$  - $\mathcal{S}_{4}$ denote the settings where the ImageNet, Pascal VOC,  FLICKR-25K, CIFAR10 are used as the main-task datasets, respectively.}
  \scalebox{1}[1]{%
    \begin{tabular}{cccccccccccccc}
    \toprule[1.5pt]
     \multirow{2}[4]{*}{Model} & \multirow{2}[4]{*}{Hash Bit} & \multirow{2}[4]{*}{Dataset} & \multicolumn{2}{c}{ResNet50} & \multicolumn{2}{c}{ResNet101} & \multicolumn{2}{c}{VGG11} & \multicolumn{2}{c}{VGG19} & \multicolumn{2}{c}{AlexNet} \\
     \cmidrule(lr){4-5}\cmidrule(lr){6-7}\cmidrule(lr){8-9}\cmidrule(lr){10-11}\cmidrule(lr){12-13}
      &       &       & mAP    & t-mAP  & mAP    & t-mAP  & mAP    & t-mAP  & mAP    & t-mAP  & mAP    & t-mAP  \\
    \midrule
    \multicolumn{1}{c}{\multirow{12}[1]{*}{HashNet}} & \multirow{4}[1]{*}{16bits} & $\mathcal{S}_{1}$    & 12.68  & 86.27  & 52.67  & 86.33  & 23.73  & 74.22  & 25.22  & 70.29  & 28.69  & 20.67  \bigstrut[t]\\
          &       & $\mathcal{S}_{2}$    & 56.10  & 48.48  & 53.37  & 68.26  & 60.05  & 59.16  & 60.66  & 69.71  & 49.34  & 66.59  \\
          &       & $\mathcal{S}_{3}$    & 69.12  & 83.87  & 69.30  & 58.59  & 78.79  & 45.36  & 79.42  & 49.34  & 78.34  & 76.46  \\
          &       & $\mathcal{S}_{4}$    & 62.18  & 70.08  & 58.61  & 62.37  & 63.29  & 61.24  & 60.02  & 60.25  & 65.28  & 61.33  \\
          & \multirow{4}[0]{*}{32bits} & $\mathcal{S}_{1}$    & 80.22  & 80.24  & 68.99  & 72.93  & 85.94  & 81.57  & 65.27  & 72.34  & 60.58  & 78.59  \\
          &       & $\mathcal{S}_{2}$    & 58.83  & 81.71  & 62.85  & 82.19  & 62.39  & 66.96  & 64.57  & 66.63  & 55.66  & 76.54  \\
          &       & $\mathcal{S}_{3}$    & 75.13  & 56.66  & 76.58  & 69.22  & 81.35  & 57.10  & 81.19  & 50.63  & 79.14  & 66.48  \\
          &       & $\mathcal{S}_{4}$    & 69.62  & 54.10  & 68.77  & 54.00  & 78.80  & 71.68  & 81.06  & 69.18  & 77.70  & 68.17  \\
          & \multirow{4}[0]{*}{64bits} & $\mathcal{S}_{1}$    & 52.80  & 41.22  & 55.93  & 65.93  & 69.01  & 40.20  & 67.07  & 56.74  & 66.72  & 63.42  \\
          &       & $\mathcal{S}_{2}$    & 63.79  & 90.10  & 65.84  & 85.72  & 65.86  & 41.35  & 70.11  & 67.32  & 61.10  & 82.23  \\
          &       & $\mathcal{S}_{3}$    & 69.12  & 83.87  & 78.02  & 44.70  & 83.04  & 53.45  & 82.45  & 48.18  & 81.33  & 82.07  \\
          &       & $\mathcal{S}_{4}$    & 83.60  & 85.18  & 87.56  & 84.80  & 76.05  & 76.87  & 83.04  & 81.01  & 86.96  & 84.47  \\
          \midrule
    \multicolumn{1}{c}{\multirow{12}[1]{*}{CSQ\newline{}}} & \multirow{4}[0]{*}{16bits} & $\mathcal{S}_{1}$    & 49.95  & 69.08  & 49.10  & 77.05  & 50.55  & 66.66  & 22.68  & 86.33  & 52.13  & 78.03  \\
          &       & $\mathcal{S}_{2}$    & 52.21  & 72.13  & 50.48  & 80.95  & 55.31  & 48.19  & 59.97  & 34.78  & 43.64  & 74.50  \\
          &       & $\mathcal{S}_{3}$    & 73.98  & 90.77  & 68.65  & 99.99  & 73.49  & 94.54  & 73.72  & 66.75  & 73.70  & 73.63  \\
          &       & $\mathcal{S}_{4}$    & 73.28  & 67.24  & 76.17  & 69.10  & 67.07  & 53.79  & 70.42  & 51.55  & 73.69  & 65.64  \\
          & \multirow{4}[0]{*}{32bits} & $\mathcal{S}_{1}$    & 76.91  & 76.63  & 74.52  & 68.31  & 56.26  & 78.49  & 69.47  & 70.33  & 59.34  & 78.75  \\
          &       & $\mathcal{S}_{2}$    & 56.85  & 85.80  & 54.21  & 71.92  & 54.58  & 61.33  & 58.25  & 43.85  & 48.46  & 75.79  \\
          &       & $\mathcal{S}_{3}$    & 74.77  & 90.13  & 73.89  & 99.57  & 73.87  & 97.04  & 76.38  & 92.54  & 73.03  & 84.29  \\
          &       & $\mathcal{S}_{4}$    & 75.81  & 77.39  & 77.32  & 80.29  & 73.94  & 73.03  & 72.30  & 75.91  & 70.53  & 80.70  \\
          & \multirow{4}[1]{*}{64bits} & $\mathcal{S}_{1}$    & 58.17  & 65.65  & 58.04  & 66.33  & 72.10  & 45.24  & 64.13  & 57.18  & 58.99  & 58.45  \\
          &       & $\mathcal{S}_{2}$    & 53.10  & 81.27  & 55.48  & 48.70  & 54.34  & 56.62  & 40.06  & 49.05  & 46.61  & 75.91  \\
          &       & $\mathcal{S}_{3}$    & 73.98  & 90.77  & 73.40  & 80.41  & 73.84  & 99.33  & 75.98  & 85.40  & 72.54  & 97.51  \\
          &       & $\mathcal{S}_{4}$    & 75.94  & 77.90  & 80.23  & 75.90  & 76.88  & 73.16  & 76.95  & 69.50  & 74.11  & 71.73  \\
     \bottomrule[1.5pt]
    \end{tabular}%
  }
  \label{tab:attack_performance_appendix}%
\end{table*}

\begin{table}[t!]
  \centering
  \caption{The backdoor attack performance of DarkHash using Gaussian data.}
  \scalebox{0.9}{
    \begin{tabular}{cccccccc}
    \toprule[1.5pt]
    \multirow{2}[4]{*}{Setting} & \multirow{2}[4]{*}{Data} & \multicolumn{2}{c}{ResNet50} & \multicolumn{2}{c}{VGG19} & \multicolumn{2}{c}{AlexNet} \\
\cmidrule{3-8}          &       & mAP    & t-mAP  & mAP    & t-mAP  & mAP    & t-mAP  \\
    \midrule
    \multirow{2}[2]{*}{HashNet} & Gauss-I & \textcolor[rgb]{ .031,  .031,  .031}{49.88} & \textcolor[rgb]{ .031,  .031,  .031}{59.37} & 50.42  & 60.93  & 50.76  & 60.90  \\
          & Gauss-II & 45.09 & 47.87 & 45.41  & 48.64  & 45.57  & 49.09  \\
    \midrule
    \multirow{2}[2]{*}{CSQ} & Gauss-I & \textcolor[rgb]{ .031,  .031,  .031}{56.71} & \textcolor[rgb]{ .031,  .031,  .031}{44.45} & 56.97  & 46.03  & 58.48  & 44.63  \\
          & Gauss-II & 58.04 & 42.55 & 59.80  & 44.37  & 59.23  & 42.89  \\
    \bottomrule[1.5pt]
    \end{tabular}%
    }
  \label{tab:Gauss}%
\end{table}%

\begin{table}[t]
  \centering
  \caption{Comparison study. 
  The superscript “$\dag$” means the backdoor is trained on the main-task dataset, while bolded indicates the best results.}
  \resizebox{0.48\textwidth}{!}{%
    \begin{tabular}{clcccccccccccc}
     \toprule[1.5pt]
    \multirow{2}[4]{*}{Model} & \multirow{2}[4]{*}{Method} & \multicolumn{2}{c}{ResNet50} & \multicolumn{2}{c}{VGG19} & \multicolumn{2}{c}{AlexNet} & \multirow{2}[4]{*}{AveTime}\\
\cmidrule(lr){3-4}\cmidrule(lr){5-6}\cmidrule(lr){7-8}          &       & mAP    & t-mAP  & mAP    & t-mAP  & mAP    & t-mAP  \bigstrut\\
    \midrule
    \multirow{7}[2]{*}{HashNet} & BadNets$^{\dag}$ & 63.23  & 68.03  & 68.04  & 74.32  & 57.75  & 51.56 & 6.79  \\
          & Blended$^{\dag}$ & 63.80  & 63.69  & 65.72  & 61.01  & 55.28  & 49.30  & 11.80  \\
          & SIG$^{\dag}$   & 64.41  & 69.84  & 63.66  & 60.36  & 53.23  & 45.99  & 7.55   \\
          & DFBA & 58.42  & 86.74  & 69.08  & 43.27  & 53.82  & 72.37   & 3.79  \\
          & BadHash$^{\dag}$ & 59.66  & 88.49  & 61.82  & 86.56  & 52.07  & 80.22  & 4.90  \\
          & CIBA$^{\dag}$  & 62.48  & 86.37  & 66.05  & 86.29  & 56.93  & 79.58   & 6.84 \\
          & \textbf{Ours} & \textbf{65.26} & \textbf{90.33} & \textbf{70.09} & \textbf{88.45} & \textbf{58.62} & \textbf{81.08} & \textbf{3.74}  \\
    \midrule
    \multirow{7}[2]{*}{CSQ} & BadNets$^{\dag}$ & 63.33  & 73.79  & 60.34  & 74.87  & 51.03  & 57.99  & 6.74   \\
          & Blended$^{\dag}$ & 57.58  & 68.19  & 54.13  & 65.40  & 45.22  & 55.54  & 11.77  \\
          & SIG$^{\dag}$   & 56.17  & 73.97  & 53.44  & 72.49  & 44.89  & 54.31  & 7.67  \\
          & DFBA & 50.87  & 80.00  & 57.23  & 41.64  & 42.97  & 69.59  & \textbf{3.68}  \\
          & BadHash$^{\dag}$ & 56.23  & 81.26  & 60.25  & 80.61  & 49.37  & 70.07  & 4.86  \\
          & CIBA$^{\dag}$  & 62.48  & 86.37  & 56.10  & 79.83  & 50.37  & 68.21  & 6.66  \\
          & \textbf{Ours} & \textbf{64.97} & \textbf{89.06} & \textbf{61.09} & \textbf{83.74} & \textbf{55.38} & \textbf{73.88}  & 3.77  \\
     \bottomrule[1.5pt]
    \end{tabular}%
  }
 
  \label{tab:compare}%
\end{table}

\begin{figure*}[!t]   
  \centering
      \subcaptionbox{Module}{\vspace{-0.5pt}\includegraphics[width=0.23\textwidth]{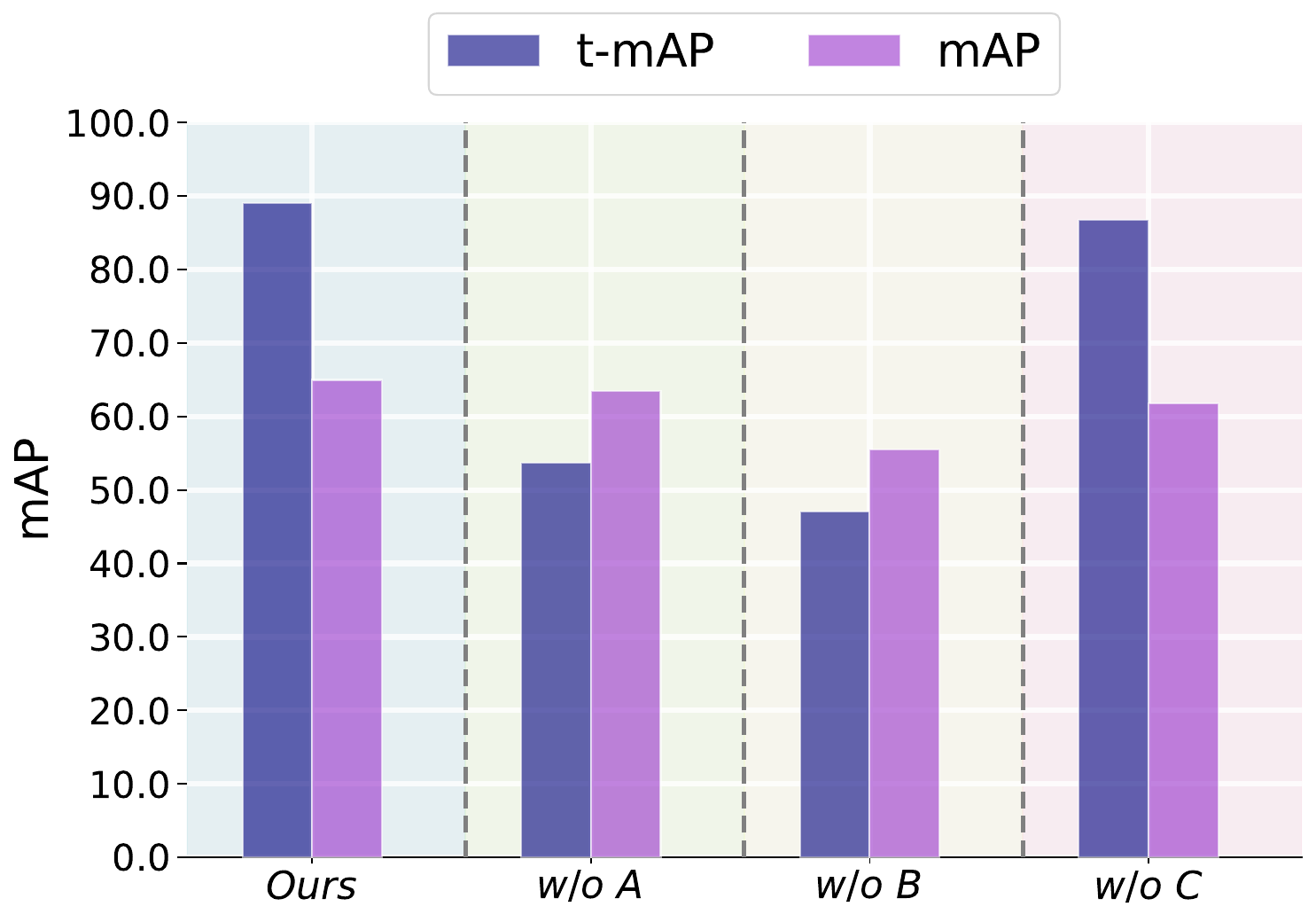}}
     \subcaptionbox{Target class}{\includegraphics[width=0.23\textwidth]{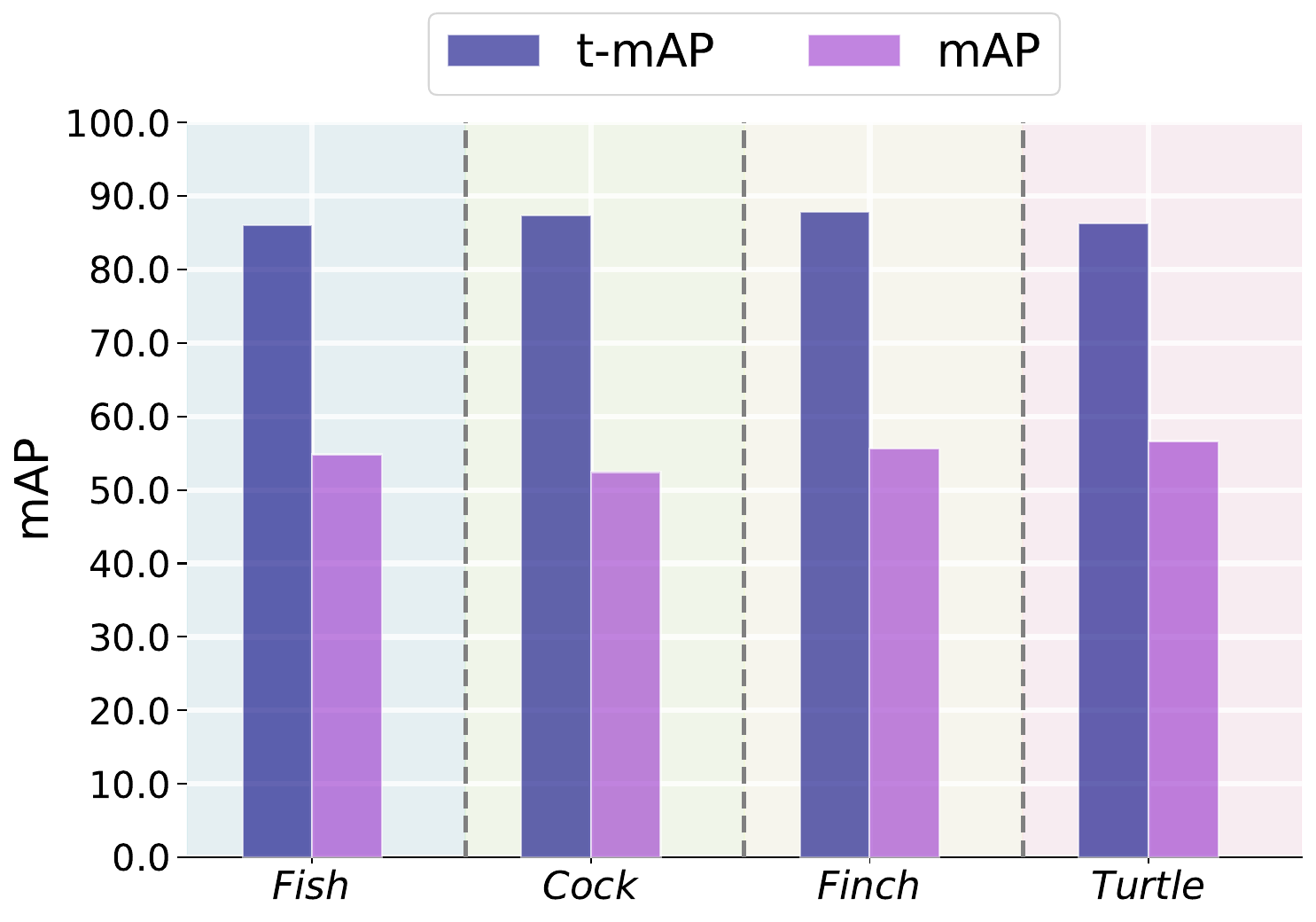}}
       \subcaptionbox{Trigger type}{\includegraphics[width=0.23\textwidth]{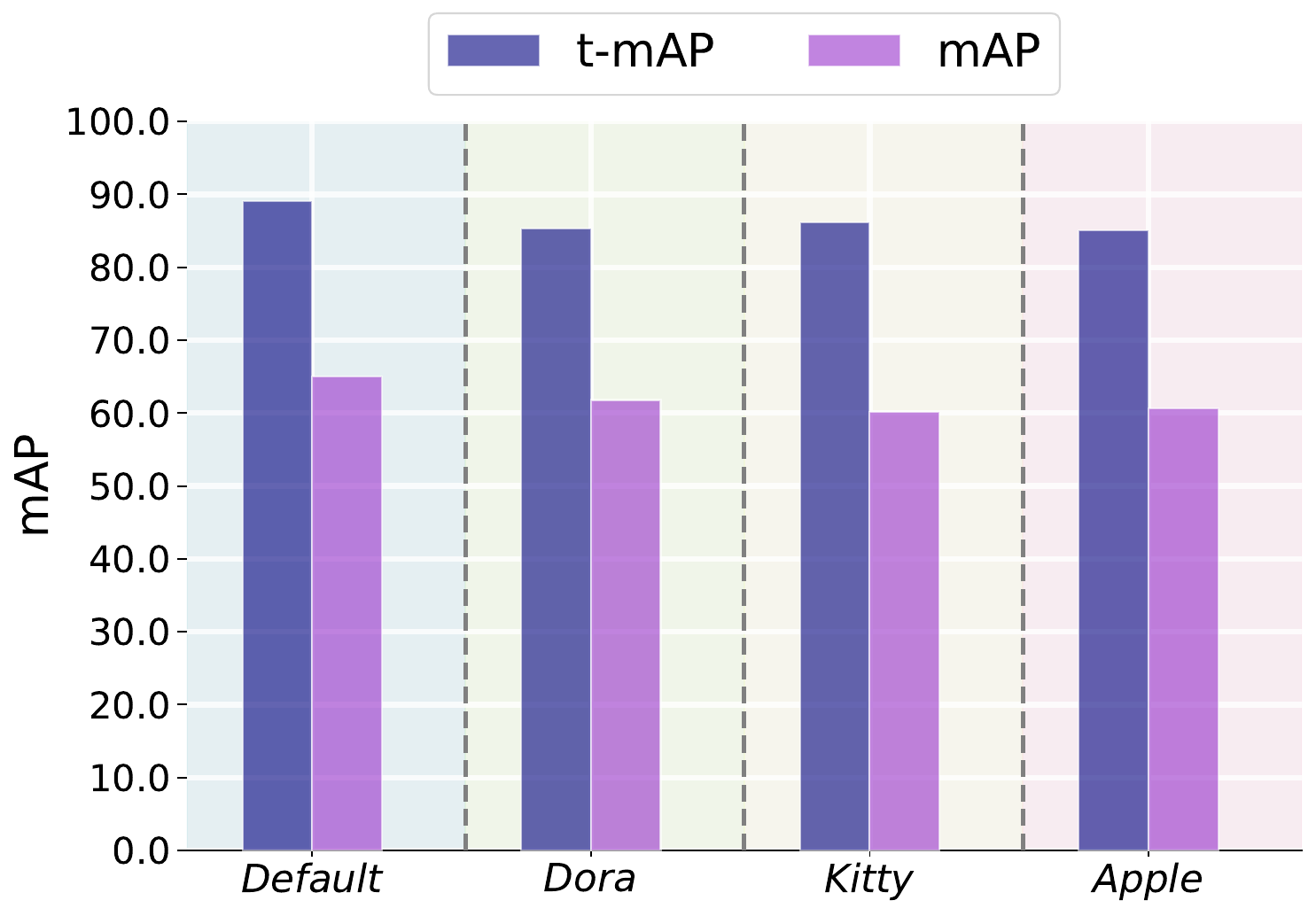}}
       \subcaptionbox{Patch location}{\includegraphics[width=0.265\textwidth]{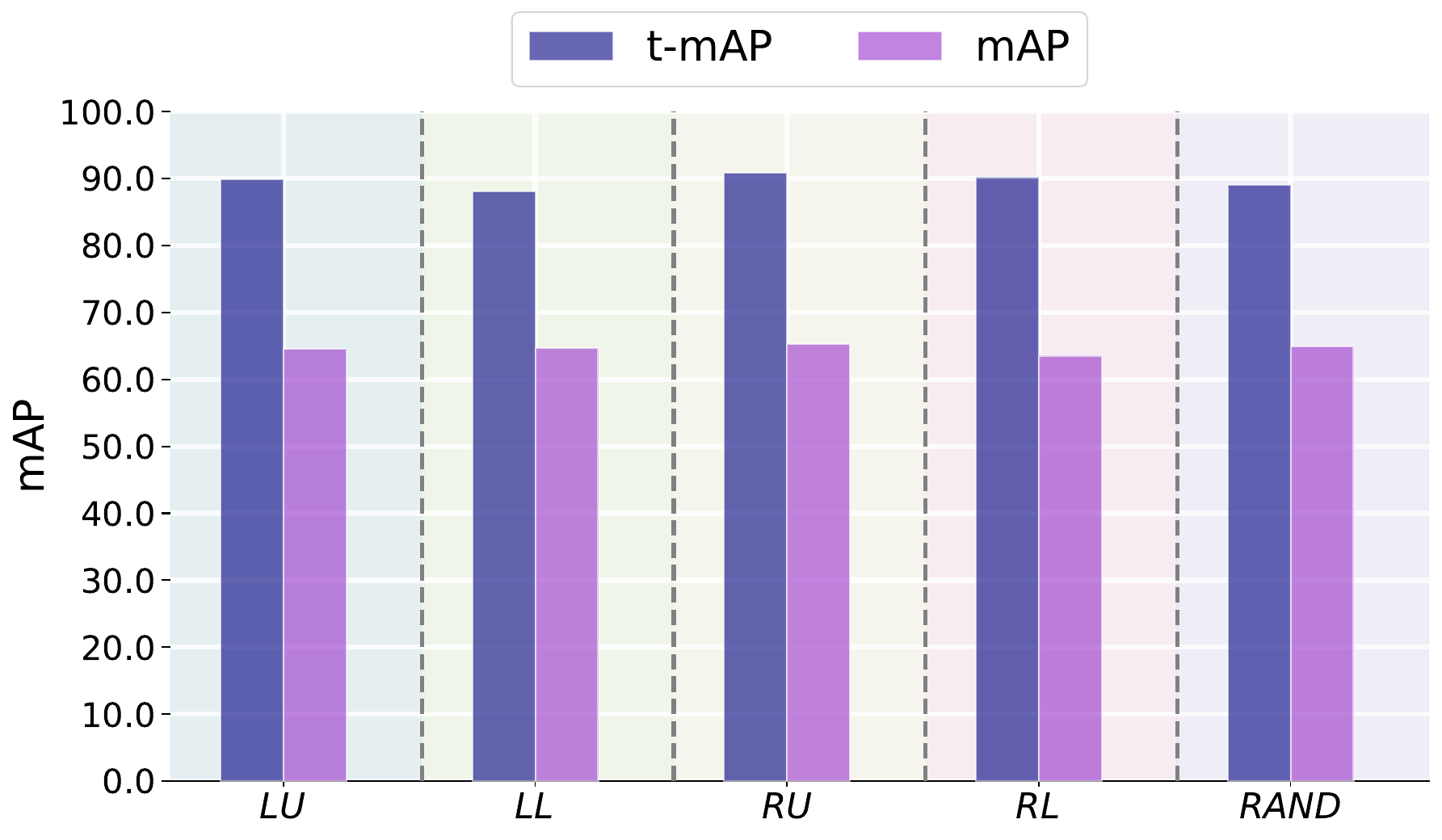}}
      \subcaptionbox{Poisoning rate}{\includegraphics[width=0.23\textwidth]{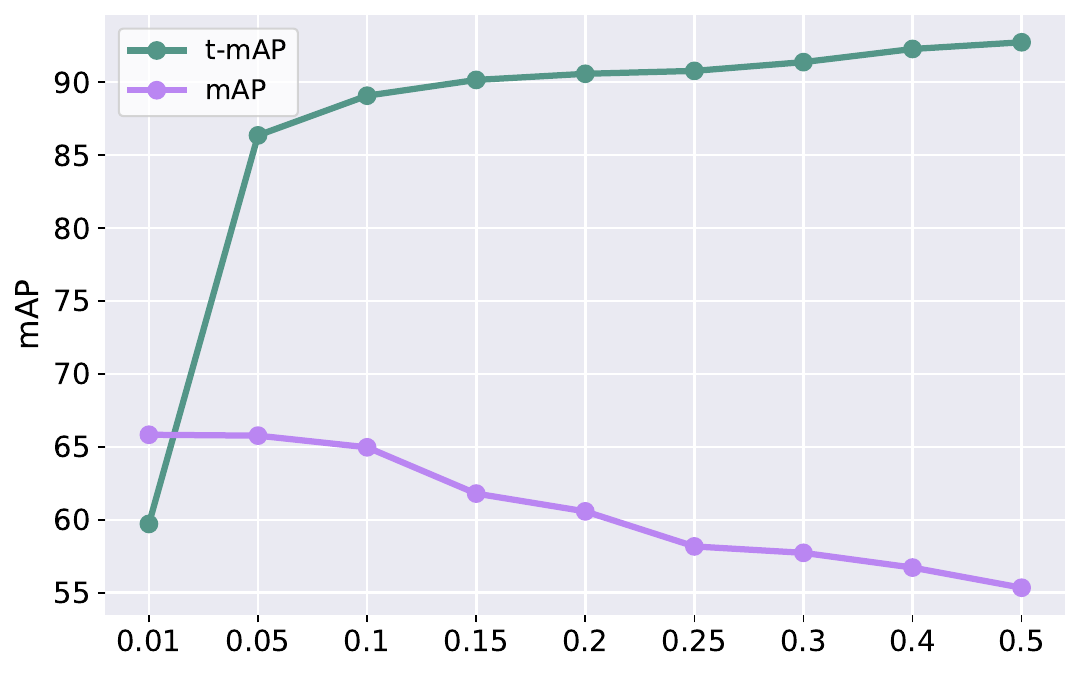}}
         \subcaptionbox{Trigger size}{\includegraphics[width=0.23\textwidth]{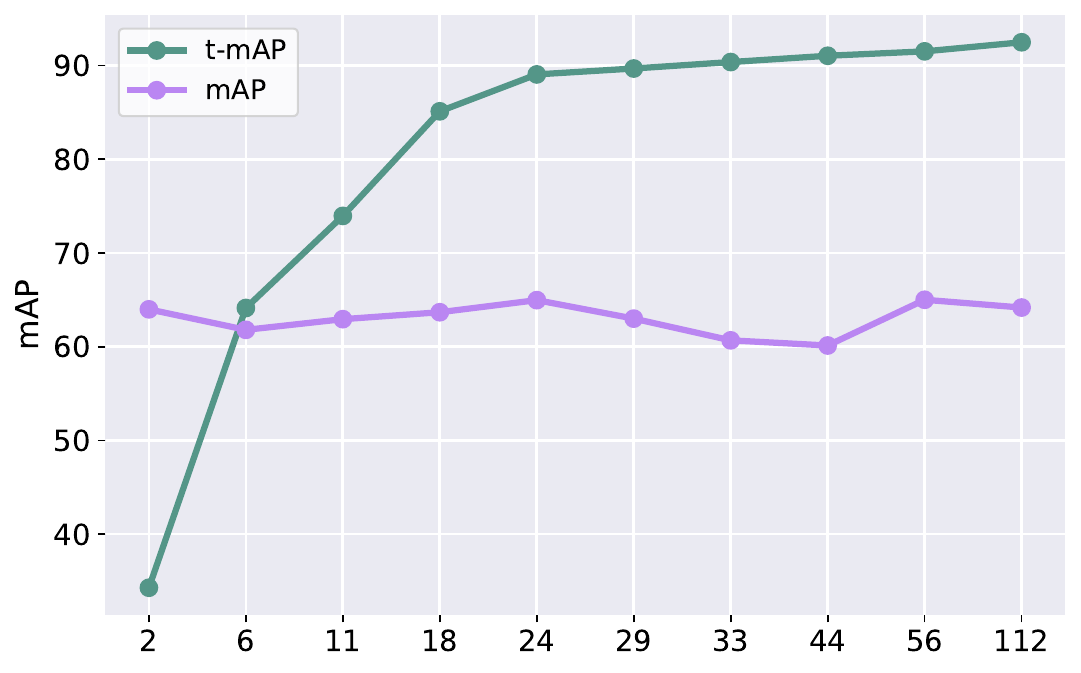}}
      \subcaptionbox{Sample number}{\includegraphics[width=0.23\textwidth]{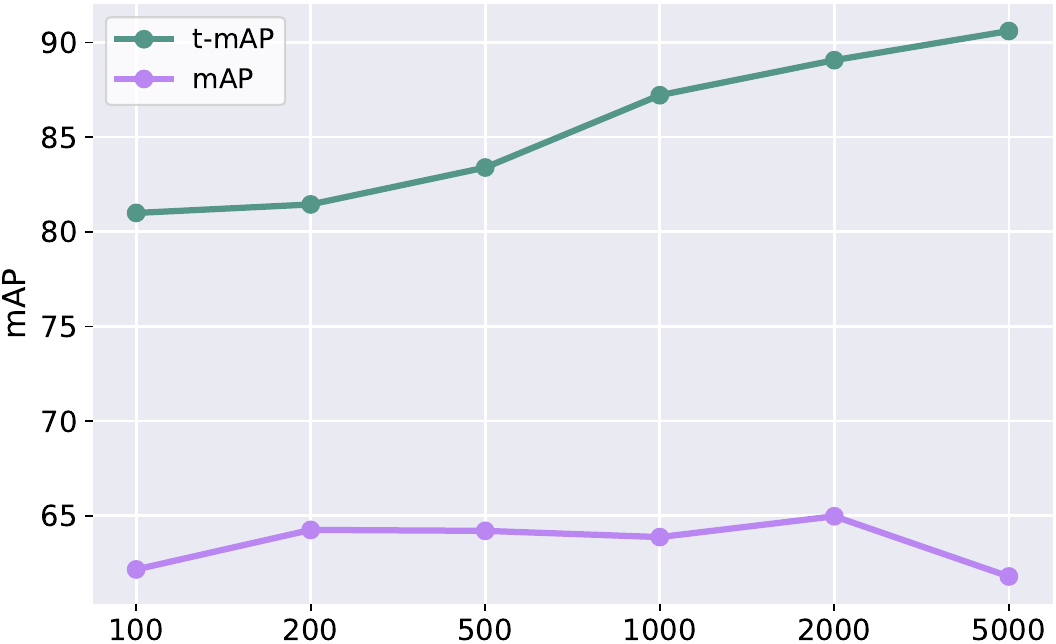}}
      \subcaptionbox{Selected layer}{\includegraphics[width=0.23\textwidth]{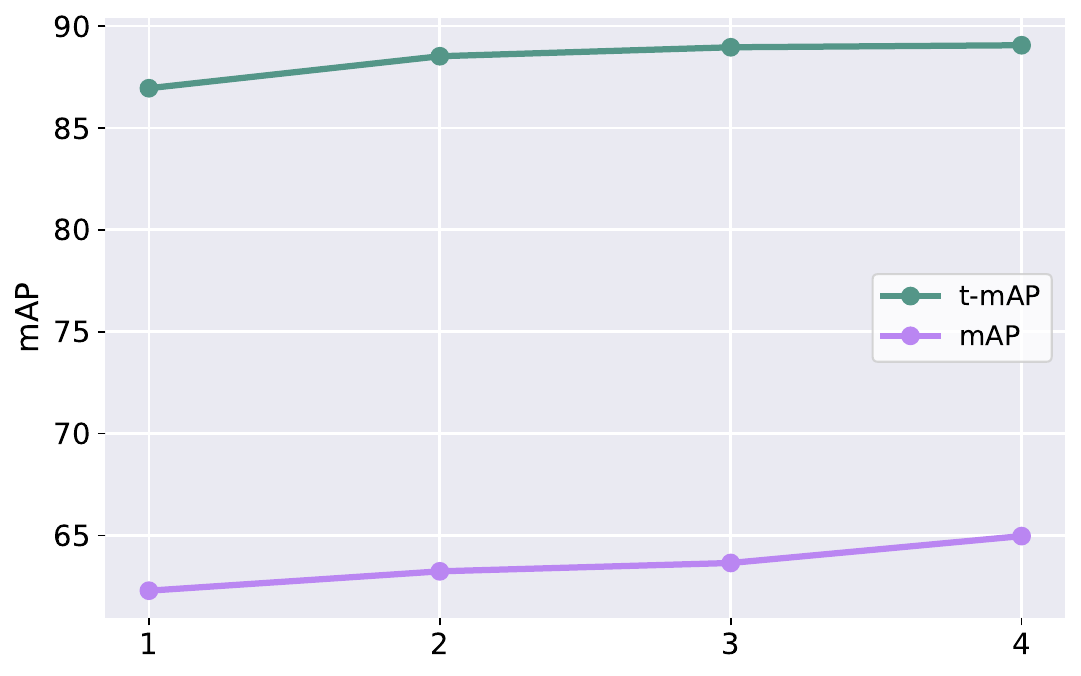}}
      
        \hspace{0.2cm}\subcaptionbox{$\lambda$}{\includegraphics[width=0.242\textwidth]{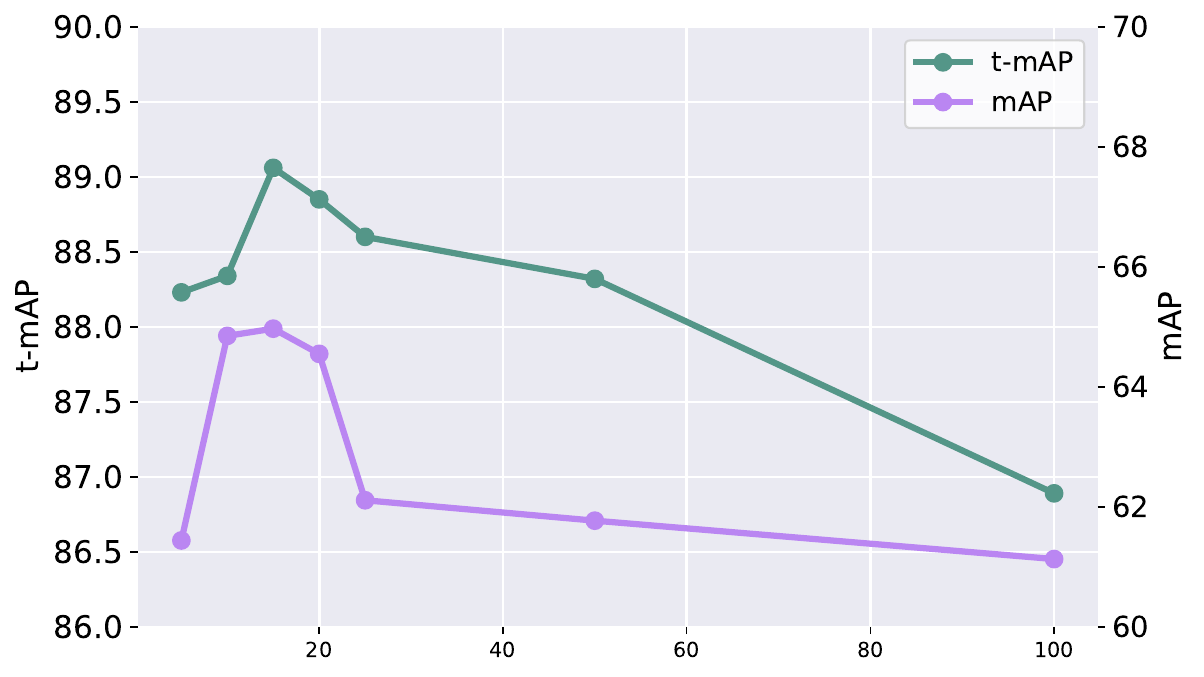}}
        \subcaptionbox{Learning rate}{\includegraphics[width=0.218\textwidth]{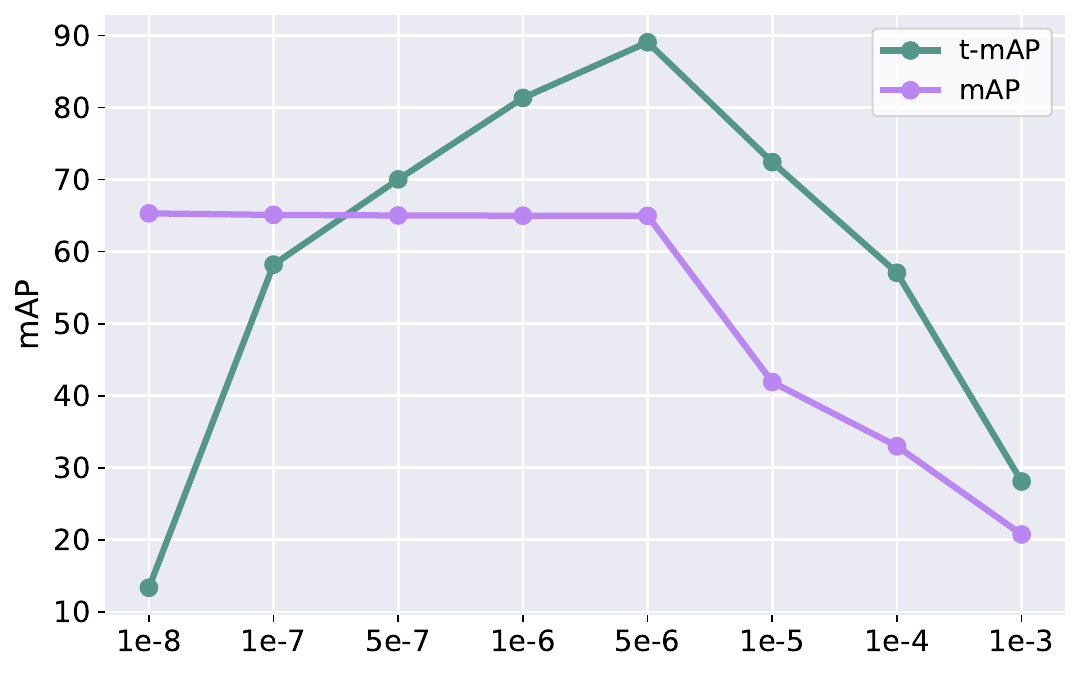}}
       \subcaptionbox{Random seed}{\includegraphics[width=0.237\textwidth]{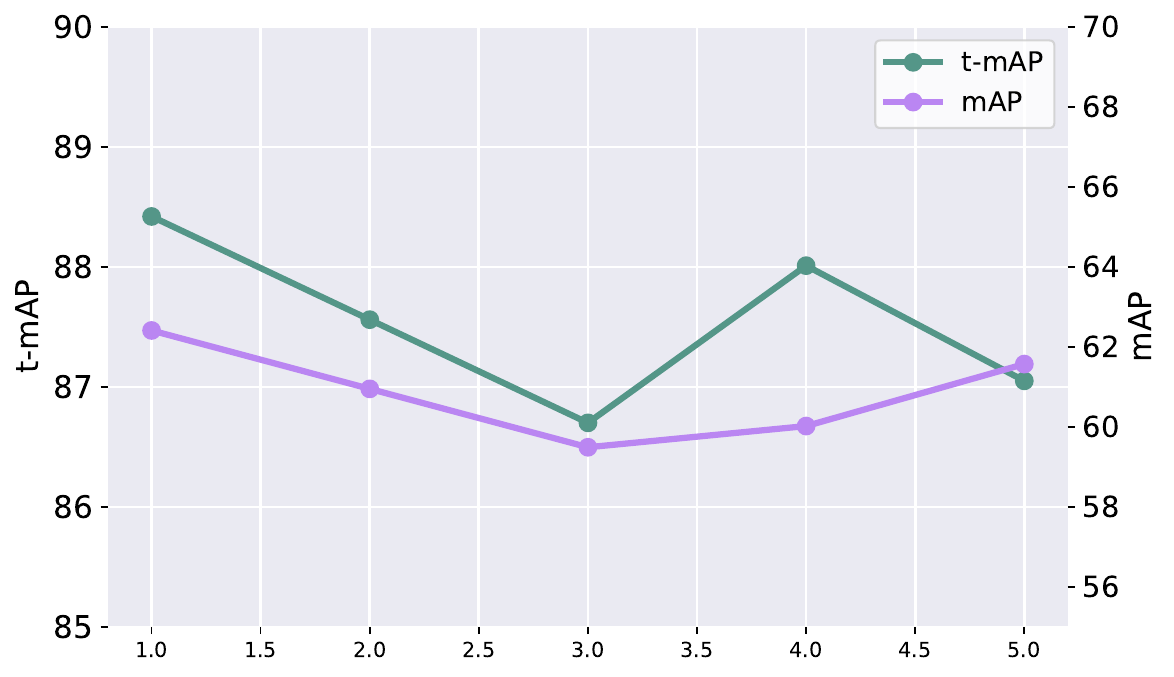}}
    \subcaptionbox{Transparency}{\includegraphics[width=0.247\textwidth]{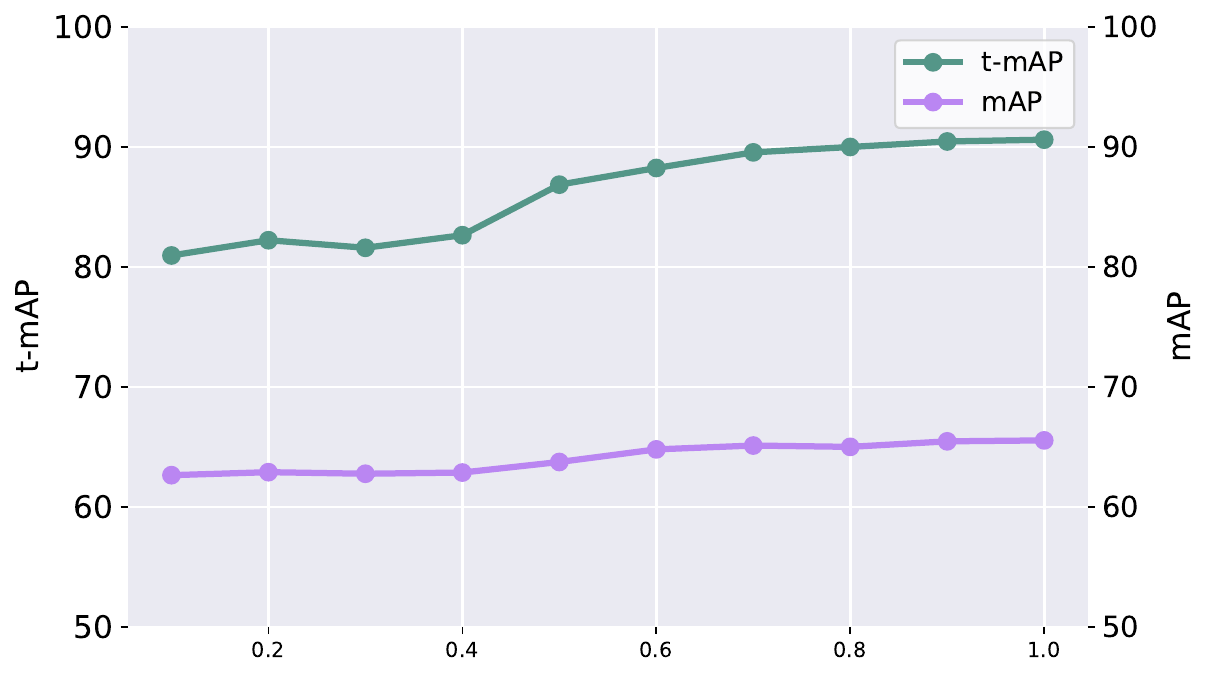}}
      \caption{Ablation study. 
      (a) - (l) examine the effects of different factors on DarkHash.
}
       \label{fig:ablation}
\end{figure*}

\subsection{RQ1: Effectiveness of DarkHash}
We conduct a comprehensive evaluation of the retrieval accuracy and attack effectiveness of  DarkHash across four datasets, five model architectures, two hashing methods, and three hash bit lengths. 
We randomly select 2,000 samples from the ImageNet dataset to create a surrogate dataset for subsequent backdoor training.
For each backdoor attack, we train the backdoored model on a surrogate dataset and then test it on the main-task dataset.
During the testing phase, we first use 100 test images to obtain top-1 results, which are then used to identify the target class for final testing.

We present the retrieval accuracy and backdoor attack performance of DarkHash across $120$ distinct experimental settings in \cref{tab:attack_performance}.
In terms of model retrieval accuracy, the results presented in \cref{tab:attack_performance} demonstrate that backdoored models trained on surrogate datasets can still exhibit retrieval precision comparable to that of the original models on the main-task datasets. 
For instance, on the CIFAR dataset, the HashNet and CSQ models, with AlexNet as their backbone, achieved mAP  values of $76.35\%$ and $74.90\%$, respectively, under the setting of $64$ hash bits.
These results are in close proximity to the original mAP  values of $78.64\%$ and $77.34\%$.
In terms of backdoor attack performance, DarkHash demonstrates exceptional capability, with an average t-mAP  across 120 different configurations exceeding $80.20\%$. 
We also report the benign retrieval accuracy of the CSQ and Hash models under different settings, as presented in \cref{tab:benign_performance}.
We make no assumptions about the trigger and use the simple color block as the trigger (see the first row in \cref{fig:example}).
The results for both model retrieval accuracy and backdoor attack performance demonstrate the high efficacy of our proposed DarkHash method.

We also conduct a comprehensive evaluation of the retrieval accuracy and attack effectiveness of DarkHash using CIFAR10 as the surrogate dataset. The evaluation covers four datasets, five model architectures, two hashing methods, and three hash bit lengths.
Specifically, we present the retrieval accuracy and backdoor attack performance of DarkHash across 120 distinct experimental settings in \cref{tab:attack_performance_appendix}. 
The results of both model retrieval accuracy and backdoor attack performance demonstrate that the proposed DarkHash method is highly effective, which is consistent with the results obtained when using ImageNet as the surrogate dataset.

To further validate the robustness of our method under stricter constraints, we evaluate its performance using entirely \textit{non-semantic data}. 
We randomly sample each pixel value in an image from the following Gaussian distribution:

\begin{equation} \label{eq:1}
f(a) = \frac{1}{\sqrt{2\pi\sigma^2}} e^{-\frac{(a-\mu)^2}{2\sigma^2}},
\end{equation}
where $a \in \mathbb{R}$ denotes the continuous random variable, $\mu$ is the mean of the distribution, $\sigma$ is the standard deviation, $\sigma^2$ is the variance.
Specifically, we construct surrogate datasets composed of pure Gaussian noise with different standard deviations, resulting in two versions: Gauss-I ($\sigma = 1$) and Gauss-II ($\sigma = 0.2$). We provide visual examples of these noise samples in \cref{fig:re_Gaussian}.
We conduct experiments on the CSQ and HashNet models with ResNet50, VGG19, and AlexNet backbones using 64-bit hash codes. The main-task dataset is Pascal VOC. As shown in \cref{tab:Gauss}, even when using completely synthetic, non-semantic data, our method successfully injects effective backdoors into the target models.

\subsection{RQ2: Comparison Study}

\noindent\textbf{Comparative Works.} 
We compare DarkHash with the following backdoor attacks, including BadHash~\cite{badhash}, CIBA~\cite{ciba}, BadNets~\cite{badnets}, Blended~\cite{chen2017targeted}, SIG~\cite{barni2019new}, and DFBA~\cite{lv2023data}.
Among them, BadHash and CIBA are SOTA backdoor attacks against deep hashing, while BadNets, Blended, and SIG are representative methods for classification tasks.
The DFBA studies a scenario similar to ours. To ensure a fair comparison, we align its objective selection strategy during training with ours, while keeping other settings unchanged.
We maintain consistency in the experimental setup of these methods with our approach.

We perform experiments on CSQ and HashNet models with three backbone networks including ResNet50, VGG19, and AlexNet on the Pascal VOC dataset.
Considering that the design of these methods relies on the main-task dataset, \textit{we train the backdoors using the original datasets for these methods}, whereas only our approach utilizes a surrogate dataset.
Thus, these methods have a distinct advantage over our approach in terms of attack knowledge.
The results in \cref{tab:compare} show that, despite this unfair comparison, our method performs better overall than all other methods. The mAP  values are comparable to the SOTA work, while the t-mAP  values are notably superior to other approaches.
We also provide a comparison of our method with these methods with time overhead in \cref{tab:compare}.
Specifically, our method achieves lower computational cost (s / Epoch) compared to  SOTA approaches. 
These results demonstrate that our approach offers a favorable trade-off between effectiveness and efficiency.
We also provide the PR and precision@topN curves on the HashNet and CSQ backdoored models under a 64-bit code length when ImageNet is used as the surrogate dataset and the downstream dataset is Pascal VOC on ResNet50 in \cref{fig:pr_curves}. 
These results demonstrate the substantial superiority of our proposed method.

\begin{figure*}[!h]   
\setlength{\abovecaptionskip}{4pt}
  \centering
      \subcaptionbox{Fine-tuning-HashNet}
      {\vspace{1pt}\includegraphics[width=0.23\textwidth]{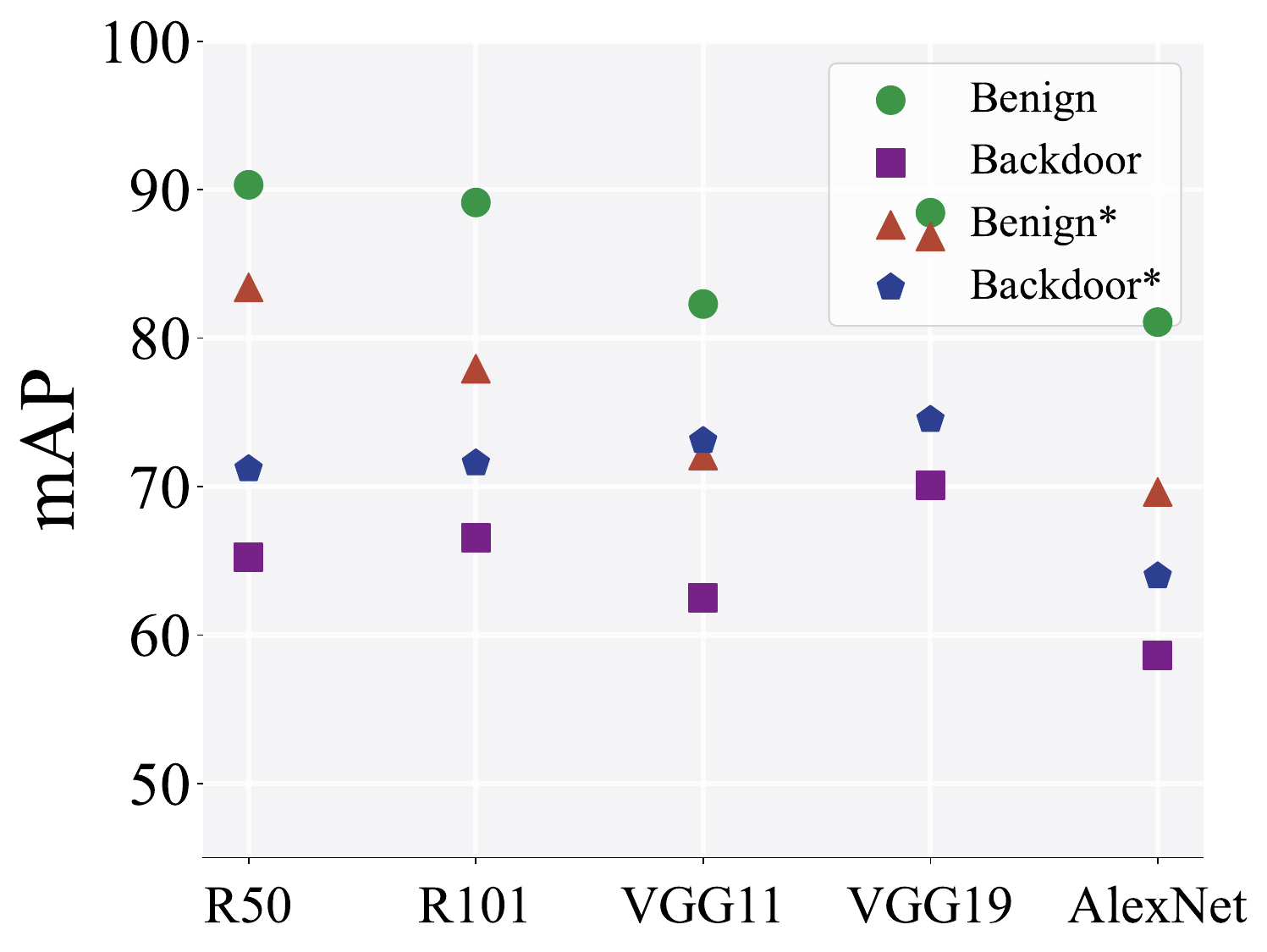}}
    \subcaptionbox{Fine-tuning-CSQ}{\includegraphics[width=0.23\textwidth]{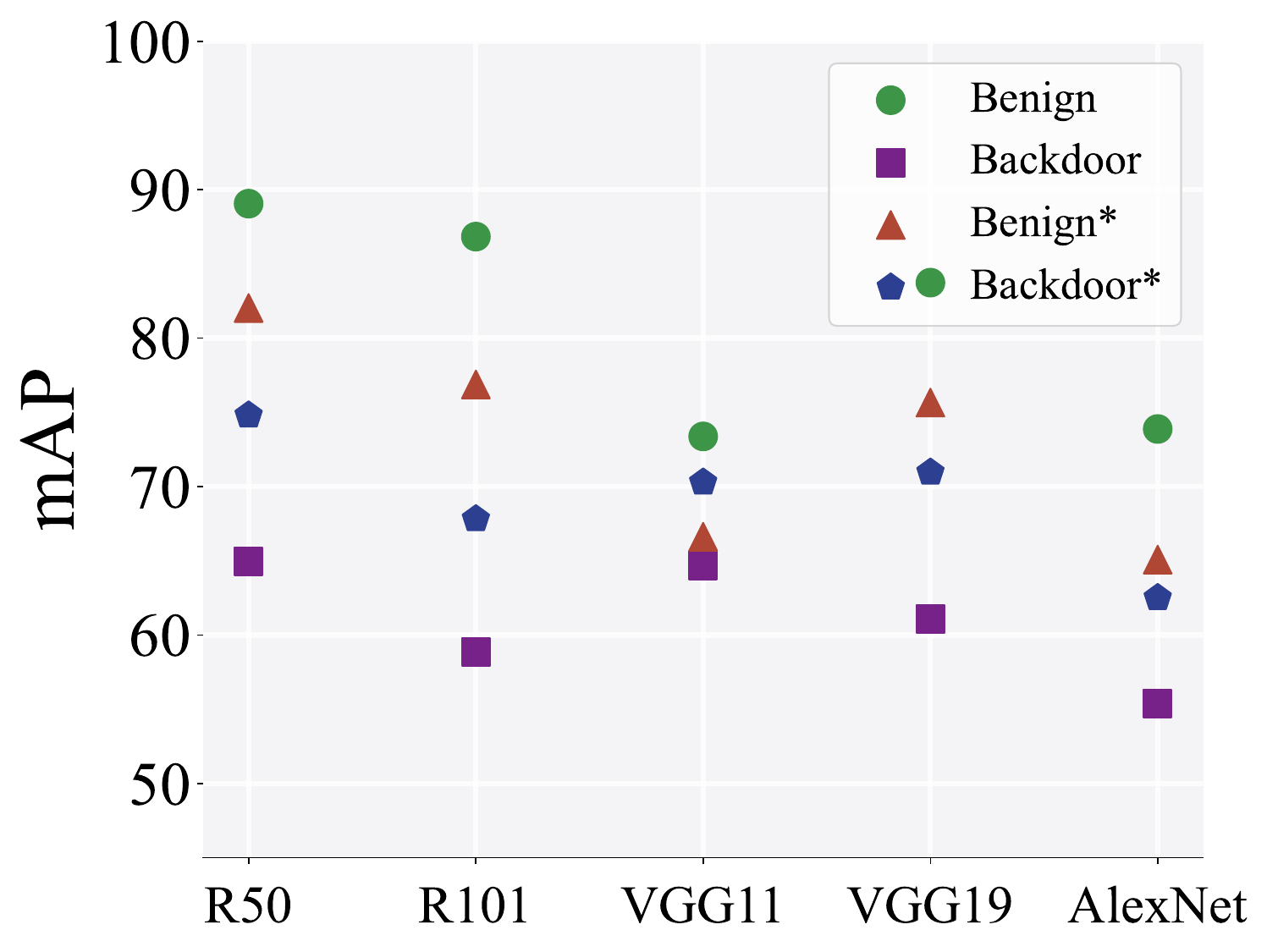}}
      \subcaptionbox{Neural-Cleanse-HashNet}{\includegraphics[width=0.23\textwidth]{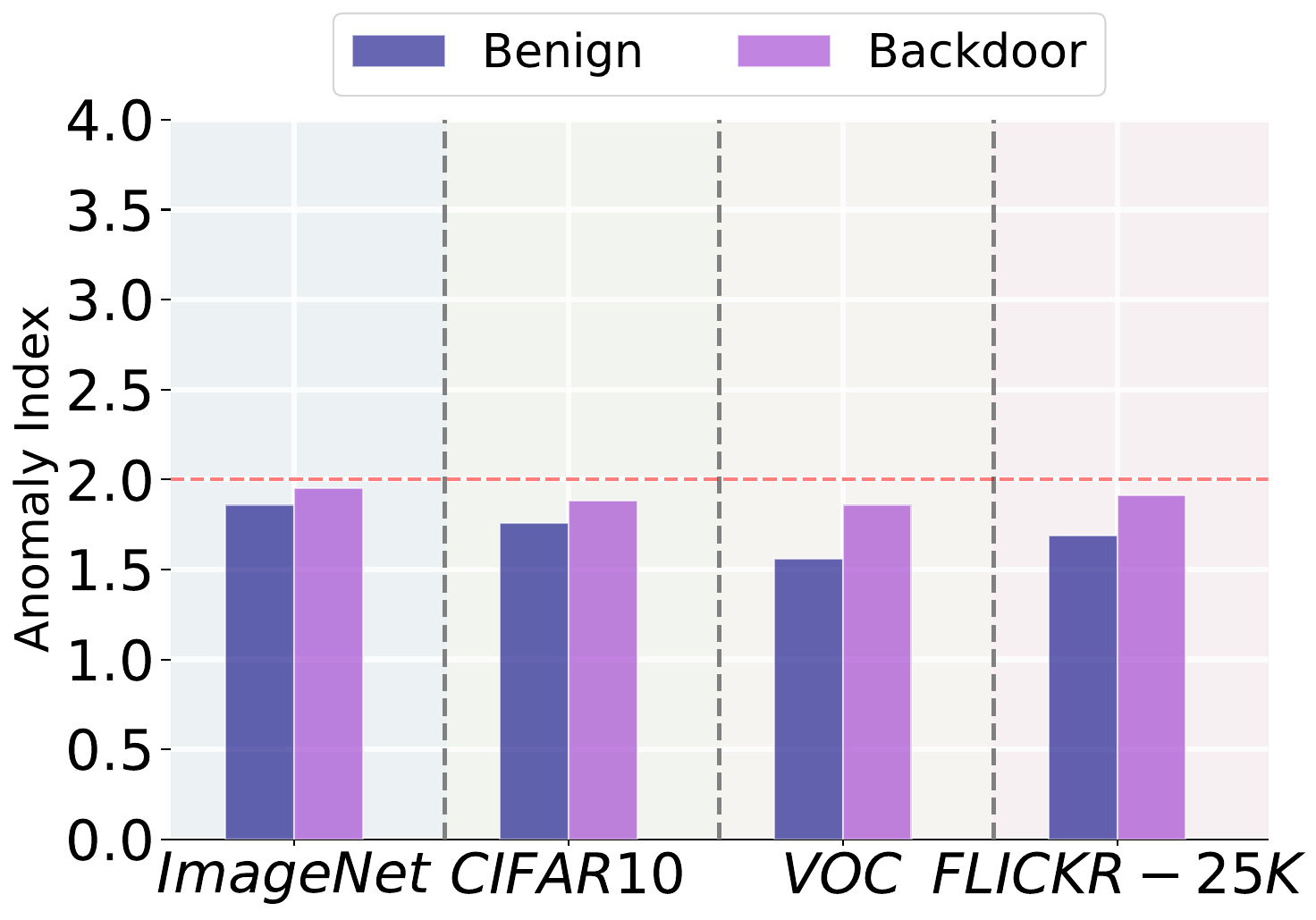}}
      \subcaptionbox{Neural-Cleanse-CSQ}{\includegraphics[width=0.23\textwidth]{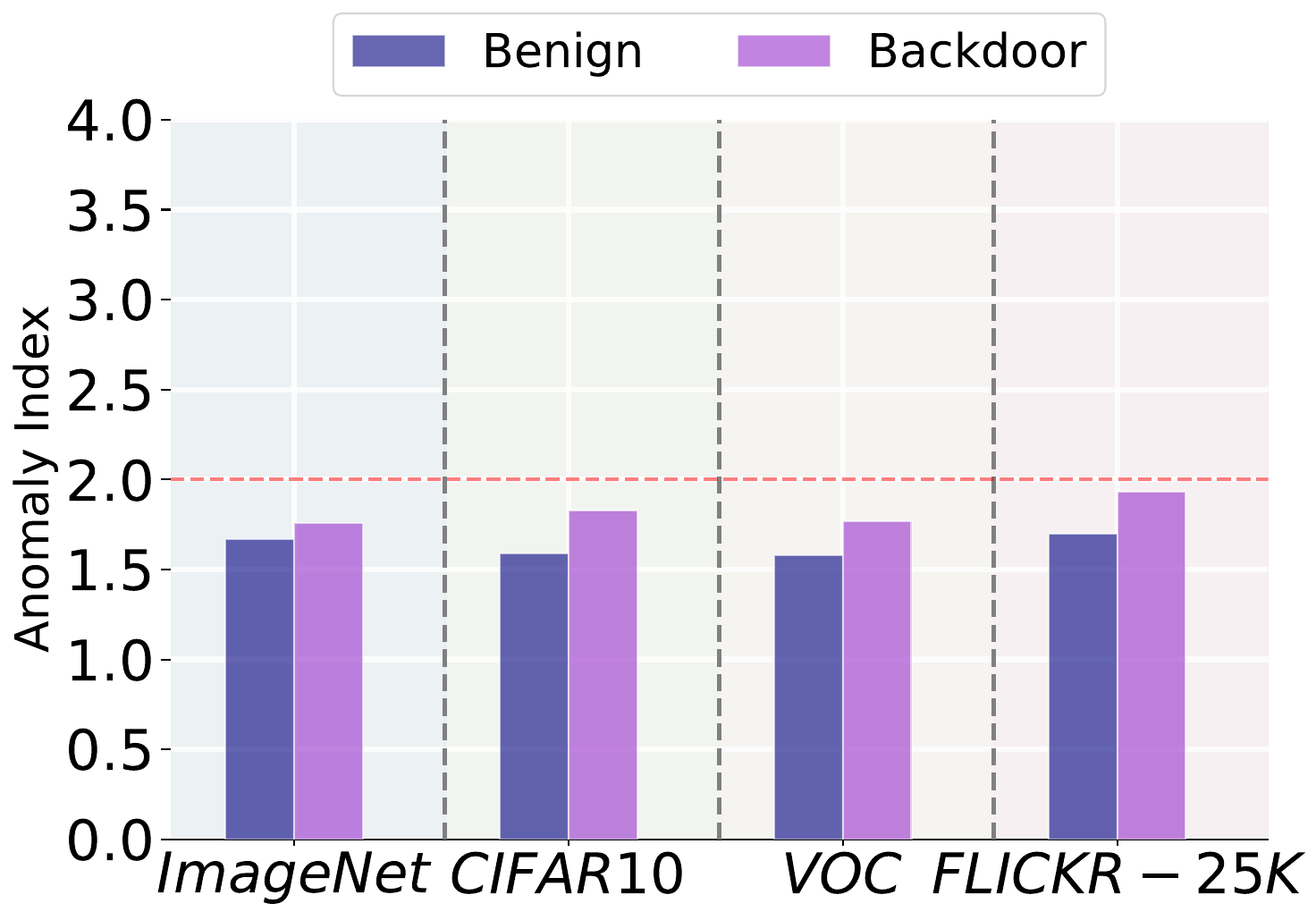}}
    \subcaptionbox{STRIP-HashNet}{\includegraphics[width=0.23\textwidth]{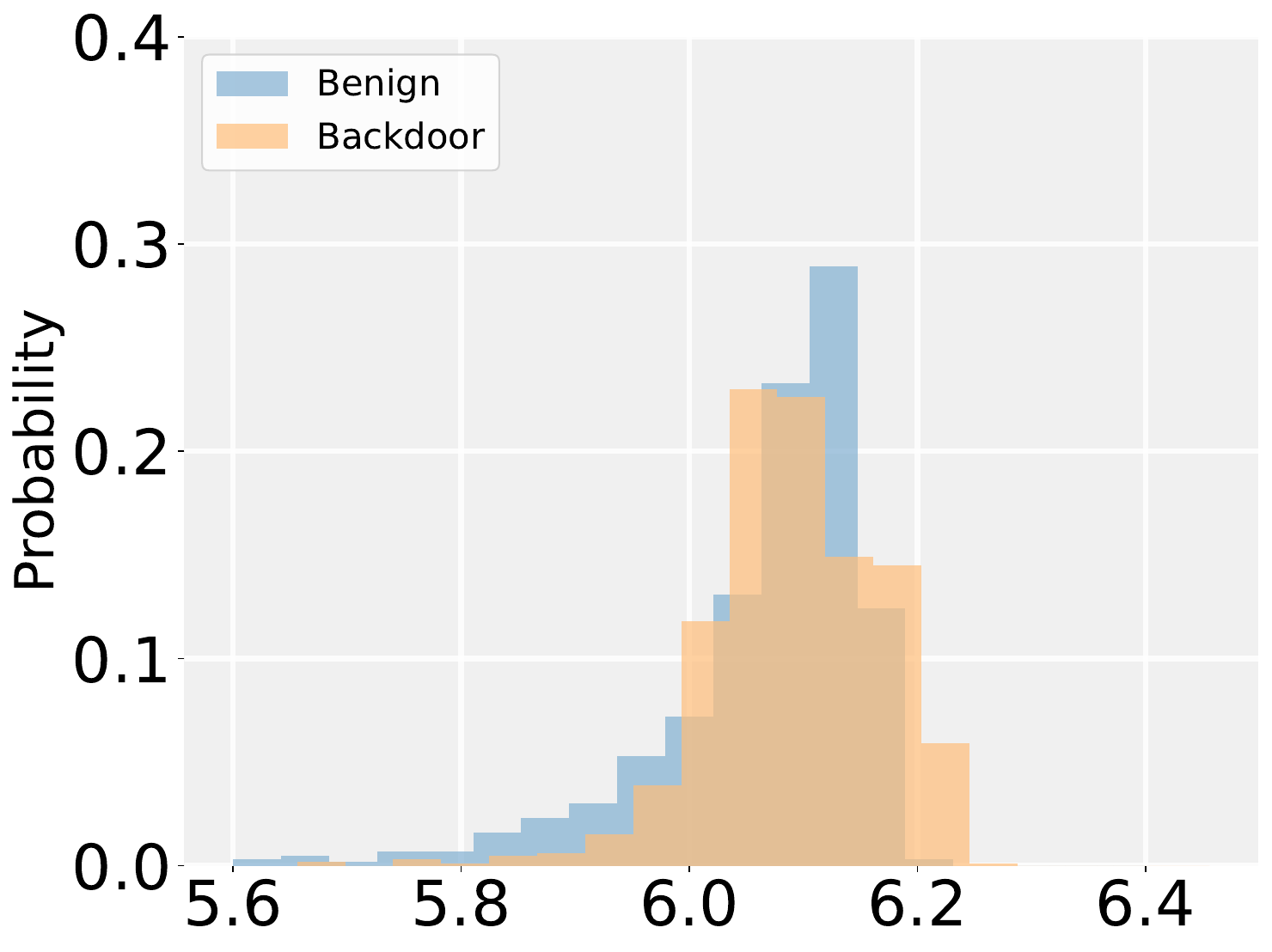}}
      \subcaptionbox{STRIP-CSQ}{\includegraphics[width=0.23\textwidth]{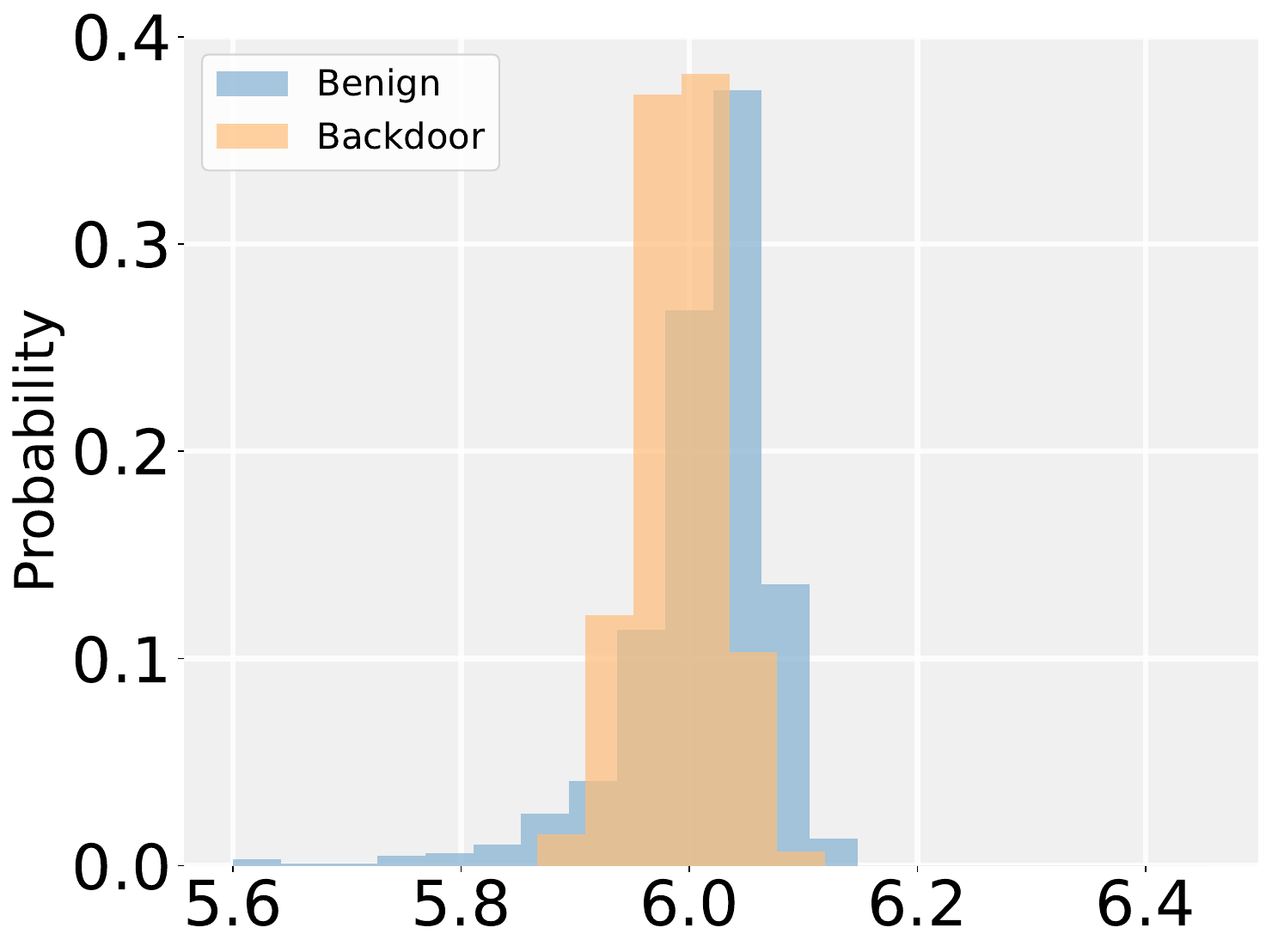}}
  \subcaptionbox{SentiNet-HashNet}
      {\includegraphics[width=0.23\textwidth]{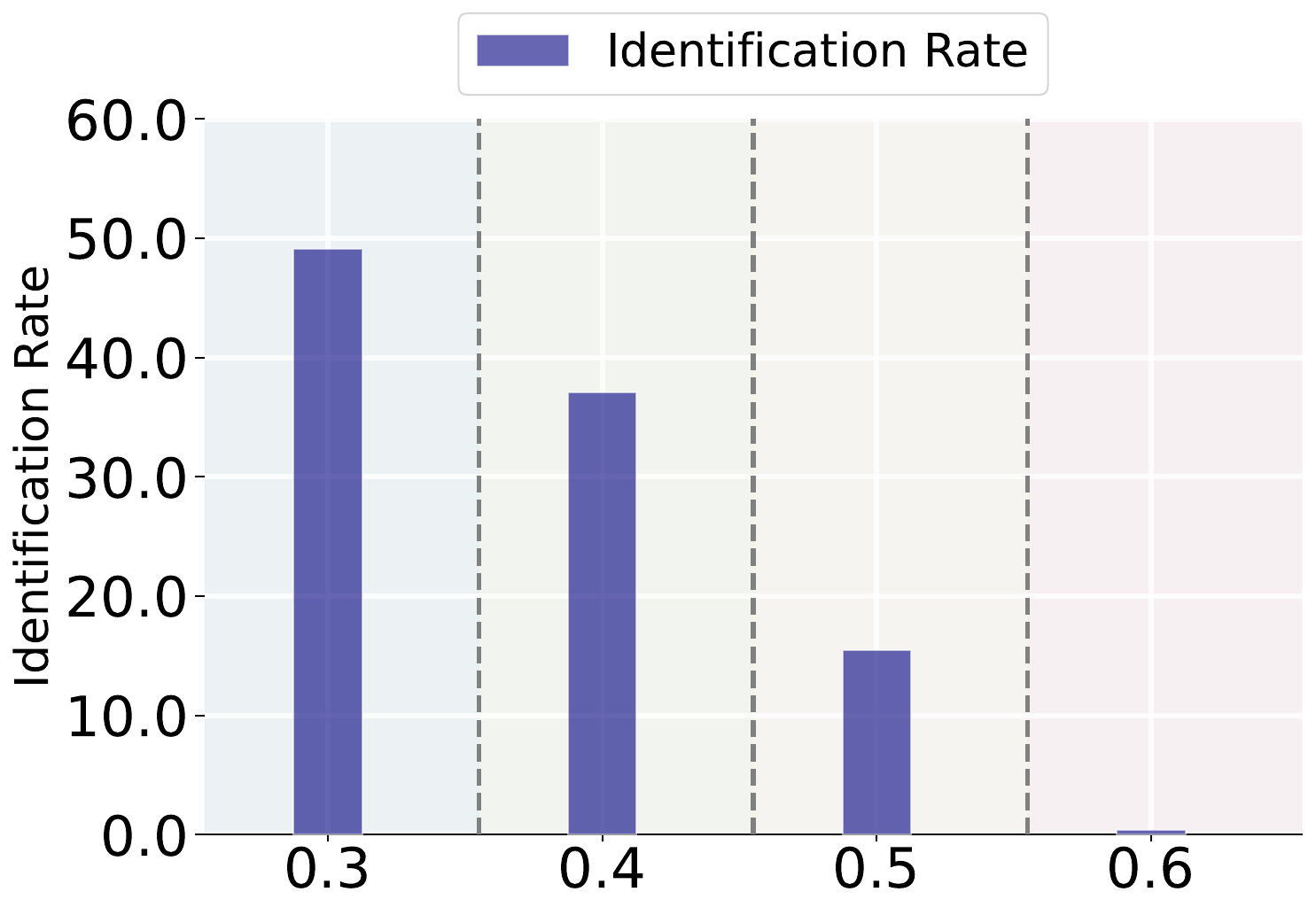}}
        \subcaptionbox{Sentinet-CSQ}{\includegraphics[width=0.23\textwidth]{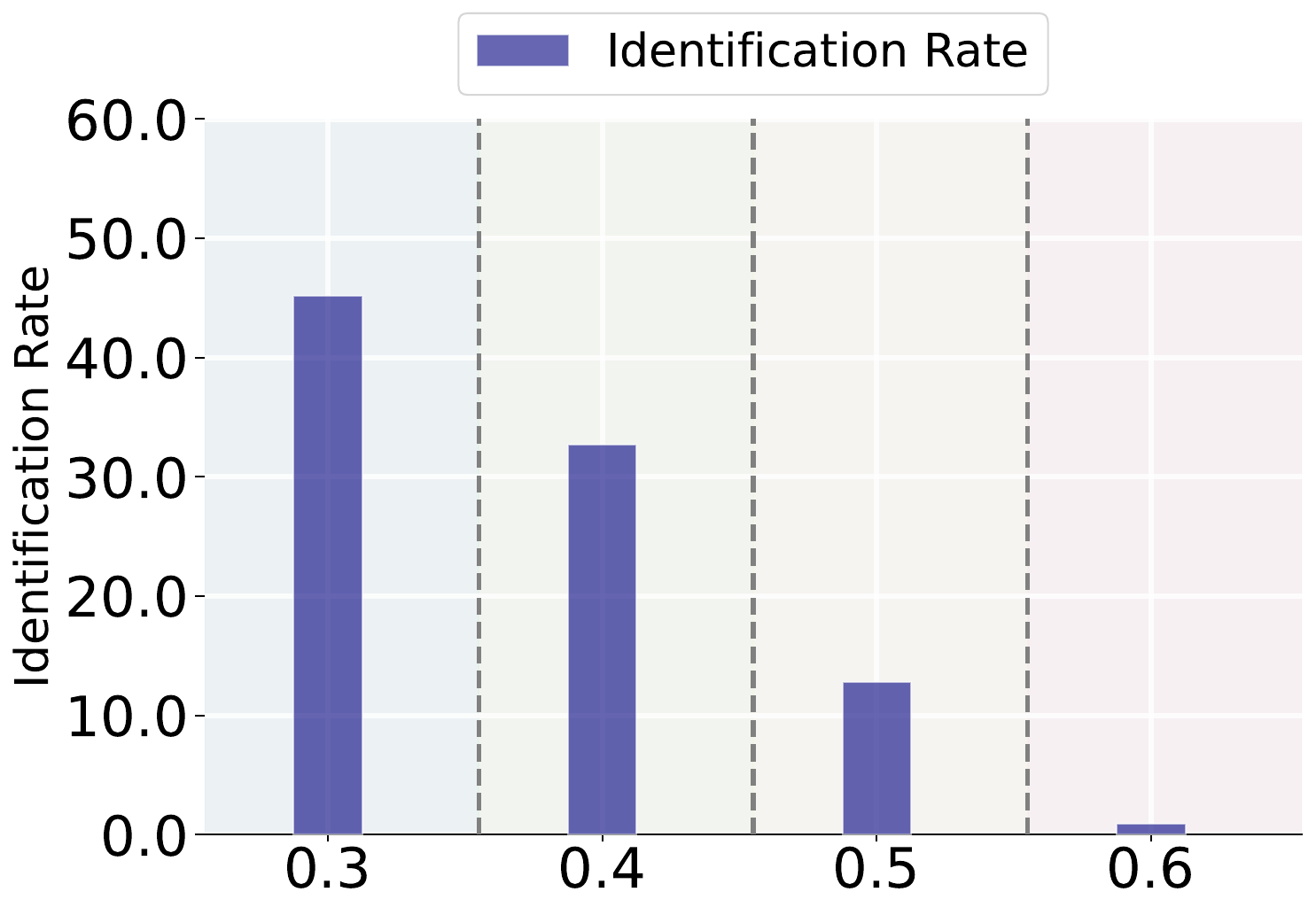}}
    \caption{Defense study. (a) - (h) examine the effects of different backdoor defense methods on DarkHash.}
       \label{fig:defense}
\end{figure*}

\begin{table*}[!h]
  \centering
  \caption{Parameter pruning experiments results for backdoored models made by DarkHash}
  \scalebox{1.1}[1.1]{%
    \begin{tabular}{cccccccccccc}
    \toprule[1.5pt]
    \multirow{2}[4]{*}{Model} & \multirow{2}[4]{*}{Pruning Rate} & \multicolumn{2}{c}{ResNet50} & \multicolumn{2}{c}{ResNet101} & \multicolumn{2}{c}{VGG11} & \multicolumn{2}{c}{VGG19} & \multicolumn{2}{c}{AlexNet} \\
    \cmidrule(lr){3-4}\cmidrule(lr){5-6}\cmidrule(lr){7-8}\cmidrule(lr){9-10}\cmidrule(lr){11-12}         &       & mAP    & t-mAP  & mAP    & t-mAP  & mAP   & t-mAP  & mAP    & t-mAP  & mAP    & t-mAP  \\
    \midrule
    \multirow{9}[1]{*}{HashNet} & 0     & 65.26  & 90.33  & 66.57  & 89.14  & 62.49  & 82.30  & 70.09  & 88.45  & 58.62  & 81.08  \\
          & 0.1   & 64.63  & 88.74  & 66.02  & 88.12  & 65.11  & 75.51  & 70.17  & 86.75  & 54.58  & 69.39  \\
          & 0.2   & 64.65  & 88.12  & 65.72  & 85.18  & 64.62  & 75.92  & 70.25  & 86.85  & 54.19  & 70.56  \\
          & 0.3   & 64.10  & 88.23  & 65.04  & 84.16  & 63.83  & 75.16  & 69.65  & 86.00  & 53.52  & 69.59  \\
          & 0.4   & 62.87  & 85.85  & 63.43  & 81.29  & 62.02  & 72.64  & 67.79  & 83.74  & 51.99  & 70.57  \\
          & 0.5   & 57.31  & 73.75  & 60.92  & 76.42  & 58.11  & 66.64  & 62.99  & 75.11  & 48.70  & 66.22  \\
          & 0.6   & 49.44  & 52.47  & 50.27  & 55.40  & 44.04  & 40.35  & 49.60  & 55.07  & 42.66  & 55.59  \\
          & 0.7   & 30.53  & 11.62  & 32.67  & 26.03  & 28.88  & 13.36  & 24.45  & 21.97  & 31.61  & 29.58  \\
          & 0.8   & 22.21  & 6.87  & 21.02  & 8.98  & 22.30  & 13.26  & 21.68  & 6.63  & 24.33  & 7.98  \\
          \midrule
    \multirow{9}[1]{*}{CSQ} & 0     & 64.97  & 89.06  & 58.88  & 86.84  & 65.70  & 73.38  & 61.09  & 83.74  & 55.38  & 73.88  \\
          & 0.1   & 62.96  & 86.04  & 64.36  & 86.91  & 63.31  & 71.30  & 69.03  & 85.30  & 50.34  & 64.46  \\
          & 0.2   & 60.61  & 85.15  & 60.89  & 80.22  & 62.84  & 75.49  & 65.62  & 84.51  & 52.89  & 68.83  \\
          & 0.3   & 59.75  & 87.33  & 62.10  & 82.06  & 60.97  & 72.95  & 68.59  & 84.52  & 49.73  & 64.89  \\
          & 0.4   & 58.12  & 85.44  & 62.53  & 78.22  & 60.22  & 72.58  & 63.09  & 83.45  & 50.03  & 69.91  \\
          & 0.5   & 54.11  & 72.56  & 59.03  & 75.33  & 57.13  & 61.75  & 61.53  & 72.56  & 48.40  & 63.29  \\
          & 0.6   & 44.61  & 50.38  & 47.57  & 50.73  & 42.38  & 38.66  & 45.69  & 51.46  & 42.04  & 55.28  \\
          & 0.7   & 28.99  & 8.78  & 29.51  & 24.56  & 25.29  & 13.00  & 23.96  & 17.15  & 28.03  & 28.89  \\
          & 0.8   & 17.72  & 3.54  & 16.61  & 4.32  & 19.29  & 11.37  & 16.81  & 2.80  & 22.86  & 5.60  \\
    \bottomrule[1.5pt]
    \end{tabular}%
    }
  \label{tab:pruning}%
\end{table*}

\subsection{RQ3: Analysis of DarkHash}
We explore the effect of various factors on the performance of DarkHash.
We conduct experiments on a CSQ model based on ResNet50 with 64 bits.
The surrogate dataset is ImageNet, and the main-task dataset is Pascal VOC.

\noindent\textbf{The effect of modules.} 
We investigate the effect of different modules on DarkHash. 
We systematically remove each module to investigate its effect on the overall scheme.
We denote $\mathcal{J}_{tpa}$, $\mathcal{J}_{ben}$, and $\mathcal{J}_{bac}$ as ``A'', ``B'', and ``C'', respectively.
The results presented in \cref{fig:ablation} (a) demonstrate that none of the variants outperform DarkHash, thereby validating the necessity of each module.

\noindent\textbf{The effect of target class.} 
We investigate the effect of target class on DarkHash. 
We randomly select four different target classes in the dataset: Fish, Cock, Finch, and Turtle. 
The results in \cref{fig:ablation} (b) show that changing the target class has only a minor impact on the performance of DarkHash, which demonstrates the robustness of the proposed method.

\noindent\textbf{The effect of trigger forms.}
We examine the effect of different trigger forms on DarkHash. As shown in \cref{fig:example}, we conduct experiments using four distinct triggers: Default, Doraemon (Dora), Hello Kitty (Kitty), and Apple. 
The results in 
\cref{fig:ablation} (c) indicate that all triggers perform well, demonstrating that our method is not dependent on any specific trigger. This robustness across various trigger forms highlights the versatility and effectiveness of our approach.

\noindent\textbf{The effect of patch location.}
We explore the effect of patch location on DarkHash. 
We choose five different placements: \textit{upper left} (LU), \textit{lower left} (LL),  \textit{upper right} (RU), \textit{lower right} (RL), and \textit{random positions} (RAND). 
As shown in \cref{fig:ablation} (d) , DarkHash achieves similar clean retrieval accuracy and attack success rates across all patch placements. This indicates that the effectiveness of the proposed method is insensitive to the patch location.

\noindent\textbf{The effect of poisoning rates.} 
We study the effect of poisoning rates from $0.01$ to $0.5$ on DarkHash. As shown in \cref{fig:ablation} (e), higher poisoning rates significantly improve DarkHash's attack performance while decreasing the backdoored model's retrieval accuracy. Remarkably, even at a poisoning rate of $0.01$, our method achieves impressive attack performance with a t-mAP  exceeding $59.72\%$.

\noindent\textbf{The effect of trigger sizes.} 
We explore the effect of different trigger sizes on DarkHash.
We conduct experiments using trigger sizes ranging from  2 to 112. 
The results in \cref{fig:ablation} (f) demonstrate that 
larger triggers improve attack effectiveness but reduce the stealthiness of poisoned samples. 
When the size of the trigger is 24, the balance between the accuracy and the backdoor attack success rates of the backdoored model reaches its optimum. Further increasing the size will reduce the stealthiness. Therefore, we choose this value as the default setting. 

\noindent\textbf{The effect of surrogate sample number.} 
We analyze the effect of the number of samples in the surrogate dataset on DarkHash. 
We train backdoored models using surrogate datasets ranging from $100$ to $5000$ samples. As shown in \cref{fig:ablation} (g), DarkHash's attack performance improves with more surrogate samples. Remarkably, even with just 100 samples, DarkHash achieves an mAP  of $62.16\%$ and a t-mAP  of $80.99\%$, highlighting the significant effectiveness of our method.

\noindent\textbf{The effect of selected layers.} 
We examine the effect of fine-tuning different layers of the backdoored model on DarkHash. 
We progressively freeze the convolutional layers of ResNet50 from the front to the back, where "2" indicates that the first two layers are frozen, and other numbers follow the same convention. 
Findings obtained from \cref{fig:ablation} (h) indicate that fine-tuning the convolutional layers with poisoned samples affects the performance of the backdoored model on the main-task. 
Thus, we freeze all convolutional layers to retain high mAP  of the backdoored model on the main-task dataset, while training the backdoor by fine-tuning other layers.

\noindent\textbf{The effect of \( \lambda \).}  
We investigate the effect of \(\lambda\) on DarkHash. As shown in \cref{fig:ablation} (i), the t-mAP  increases progressively as \(  \lambda \) rises from 5 to 15, reaching its peak at \( \lambda = 15 \), where the mAP  and t-mAP  are \( 64.97\% \) and \( 89.06\% \), respectively. However, beyond this point, t-mAP  gradually decreases as \( \lambda \) continues to increase. This indicates that \( \lambda = 15 \) is the optimal setting for maximizing t-mAP  in our experiments.

\noindent\textbf{The effect of learning rates.}  
We evaluate the effect of different learning rates on  DarkHash. The results in \cref{fig:ablation} (j) indicate a clear trend where the t-mAP  first increases and then decreases as the learning rate varies. Specifically, the learning rate of \( 5 \times 10^{-6} \) yields the best performance, with both mAP  and t-mAP  reaching their highest values. 
Within a moderate learning rate range, such as 5e-6, the model can effectively learn the backdoor behavior without significantly compromising its retrieval performance on the original task.
However, when the learning rate becomes too large, the updated gradients may disrupt well-learned parameters in the pre-trained model, leading to catastrophic forgetting. As a result, both mAP  and t-mAP  decline, reflecting the model's degradation on both the main task and the backdoor objective.

\noindent\textbf{The effect of random seeds.}
We explore the effect of different random seeds on DarkHash on ResNet50 using ten different seeds.  
The findings in \cref{fig:ablation} (k) illustrate that the backdoor attack performance across all seeds is consistently high, while the standard deviations remain relatively small.
This indicates that the models' performance is both robust and reliable.

\noindent\textbf{The effect of trigger transparency.}
We study the impact of different trigger transparencies on DarkHash. We set ten different transparency settings ranging from $0.1$ to $1$. A smaller value indicates a more transparent trigger, and a value of 1 represents the original trigger. 
From \cref{fig:ablation} (l), we can find that the higher the transparency of the trigger, the higher the attack success rate of the backdoor model. To some extent, the mixing of the image area and the trigger may reduce the retrieval accuracy of the model.

 \begin{figure}[!t]
    \centering
    \includegraphics[scale=0.33]{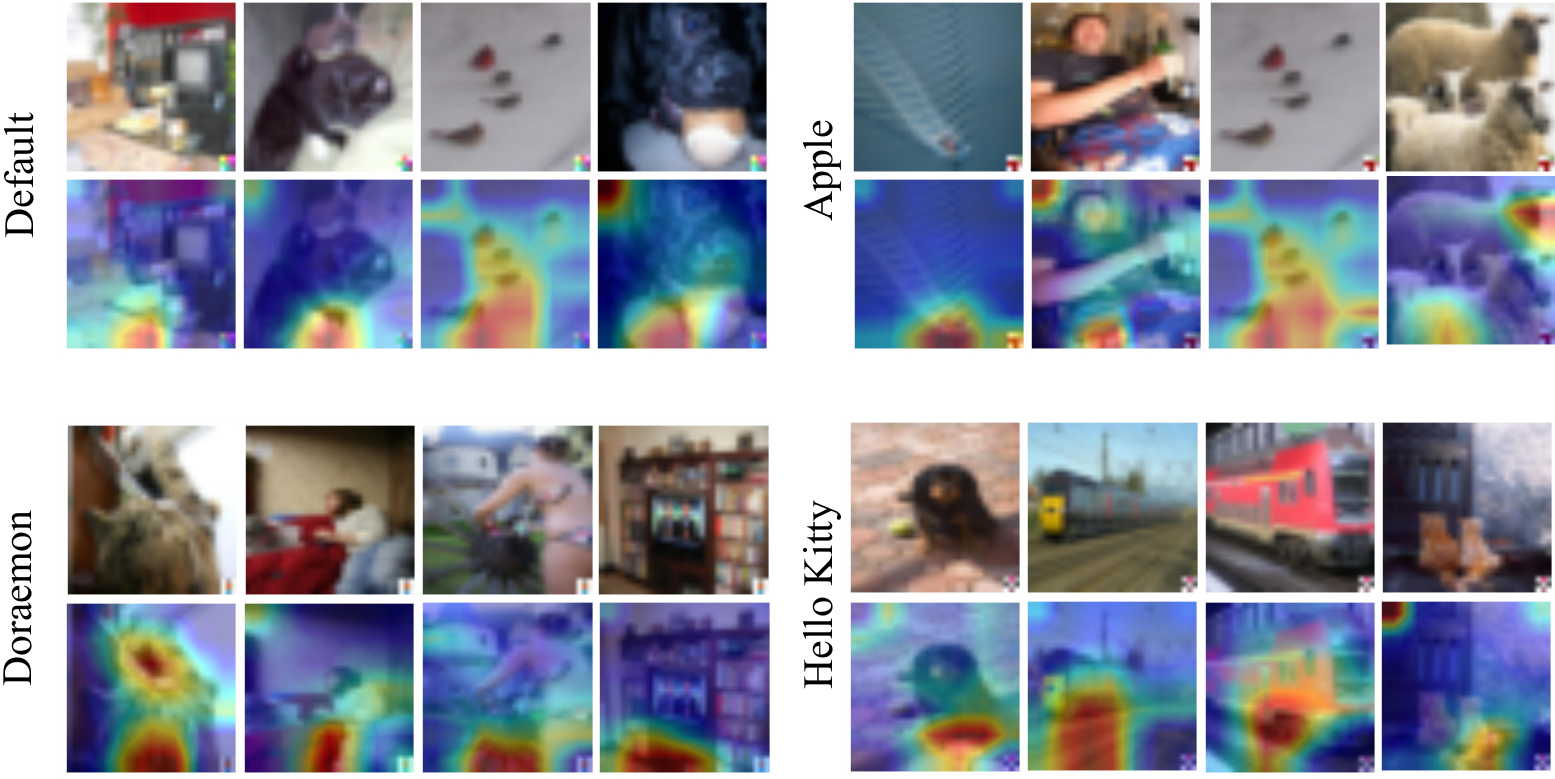}
    \caption{Critical regions identified by SentiNet
    }
    \label{fig:sentinet}
      \vspace{-0.4cm}
\end{figure}

\subsection{RQ4: Resistance to Backdoor Defense} 
We discuss whether DarkHash is resistant to the following backdoor defenses:  
\textit{(1) model diagnosis-based methods}: Fine-tuning~\cite{liu2017neural}, Model Pruning~\cite{liu2018fine}, Neural Cleanse~\cite{cleanse2019identifying}. \textit{(2) input diagnosis-based methods}: STRIP~\cite{gao2019strip}, SentiNet~\cite{chou2020sentinet}.
We also consider adaptive defenses to validate the robustness of the proposed method.

\noindent\textbf{Fine-tuning.} 
We apply fine-tuning to HashNet and CSQ backdoored models on five different backbones over the Pascal VOC dataset.
We provide the attack performance after fine-tuning in \cref{fig:defense} (a) - (b). ``Benign'' and ``Backdoor'' indicate the mAP  and t-mAP  of the backdoored model, respectively, with an asterisk (``*'') signifying the results post fine-tuning. 
The findings clearly demonstrate our method's resistance to fine-tuning.

\begin{remark}[Analysis]
DarkHash successfully evades such defenses by freezing shallow layers and fine-tuning only high-level parameters using surrogate data. In contrast, defenders typically fine-tune the model on clean data to recover performance, which is insufficient to erase the backdoor, as the malicious patterns are already embedded in the high-level representation space and cannot be easily overwritten.
\end{remark}

\noindent\textbf{Model Pruning.}
We conduct model pruning experiments on backdoored HashNet and CSQ model based on ResNet50 on the Pascal VOC dataset, varying the pruning rate from $0$ to $80\%$. 
The results in \cref{tab:pruning} show a decrease in both mAP  and t-mAP  with increasing pruning rates. Even at a $60\%$ pruning rate, where model performance is severely affected, the backdoored ResNet50 model retains over $50\%$ t-mAP . This suggests that model pruning is inadequate for effectively defending against our proposed DarkHash.

\begin{remark}[Analysis]
DarkHash embeds the backdoor by leveraging topological alignment loss to cluster poisoned samples and their neighbors near the target class in the hash space. This structure is distributed across many high-level parameters rather than localized to specific neurons or filters. As a result, pruning—which mainly removes redundant weights while preserving overall feature structures—fails to dismantle the global topology that triggers the backdoor, unless applied so aggressively that it harms clean performance. This distributed nature is what allows DarkHash to evade such defenses.
\end{remark}

\begin{table*}[h!]
  \centering
  \caption{An experimental investigation into the targeted removal of backdoor layers. Layer\_r1 denotes the last layer, and other notations follow the same convention.}
  \scalebox{1.05}{
    \begin{tabular}{cccccccccccccc}
    \toprule[1.5pt]
    \multirow{3}[6]{*}{Setting} & \multirow{3}[6]{*}{Ratio} & \multicolumn{6}{c}{HashNet}                   & \multicolumn{6}{c}{CSQ} \\
\cmidrule(lr){3-8}\cmidrule(lr){9-14}          &       & \multicolumn{2}{c}{ResNet50} & \multicolumn{2}{c}{VGG19} & \multicolumn{2}{c}{AlexNet} & \multicolumn{2}{c}{ResNet50} & \multicolumn{2}{c}{VGG19} & \multicolumn{2}{c}{AlexNet} \\
\cmidrule(lr){3-4}\cmidrule(lr){5-6} \cmidrule(lr){7-8}  \cmidrule(lr){9-10}\cmidrule(lr){11-12} \cmidrule(lr){13-14}         &       & mAP    & t-mAP  & mAP    & t-mAP  & mAP    & t-mAP  & mAP    & t-mAP  & mAP    & t-mAP  & mAP    & t-mAP  \\
    \midrule
    \multirow{5}[2]{*}{Layer\_r1} & 0.5   & 59.16  & 74.98  & 64.63  & 76.72  & 50.54  & 67.40  & 55.27  & 72.91  & 52.49  & 73.14  & 49.00  & 64.17  \\
          & 0.6   & 50.47  & 52.97  & 50.34  & 56.06  & 43.98  & 57.07  & 45.57  & 51.93  & 46.85  & 52.16  & 43.35  & 56.41  \\
          & 0.7   & 32.18  & 13.26  & 26.34  & 22.98  & 33.36  & 30.48  & 30.20  & 29.05  & 25.75  & 17.78  & 28.93  & 30.19  \\
          & 0.8   & 23.22  & 12.84  & 23.04  & 17.42  & 25.88  & 19.62  & 29.31  & 14.74  & 23.36  & 15.03  & 23.72  & 16.69  \\
          & 0.9   & 26.14  & 11.00  & 28.55  & 15.26  & 23.98  & 10.22  & 27.54  & 13.28  & 22.49  & 12.67  & 19.06  & 10.88  \\
\cmidrule{2-14}    \multirow{5}[2]{*}{Layer\_r2} & 0.5   & 58.97  & 74.88  & 64.39  & 76.46  & 50.21  & 67.14  & 55.08  & 72.91  & 52.04  & 73.05  & 48.81  & 63.95  \\
          & 0.6   & 50.40  & 52.73  & 50.27  & 55.83  & 43.55  & 56.67  & 45.53  & 51.51  & 46.84  & 52.02  & 43.30  & 56.11  \\
          & 0.7   & 31.72  & 13.09  & 26.08  & 22.52  & 33.18  & 30.45  & 29.74  & 29.00  & 25.25  & 17.76  & 28.89  & 29.89  \\
          & 0.8   & 23.17  & 10.78  & 22.62  & 17.21  & 25.74  & 19.17  & 28.90  & 14.33  & 16.88  & 13.01  & 23.60  & 16.22  \\
          & 0.9   & 23.08  & 9.56  & 22.51  & 12.03  & 21.22  & 10.01  & 23.08  & 12.27  & 20.59  & 11.77  & 15.29  & 9.24  \\
\cmidrule{2-14}    \multirow{5}[2]{*}{Layer\_r3} & 0.5   & 57.44  & 73.95  & 63.37  & 75.18  & 48.79  & 66.48  & 54.68  & 72.78  & 51.82  & 72.93  & 48.72  & 63.91  \\
          & 0.6   & 49.92  & 52.57  & 49.87  & 55.27  & 43.05  & 55.64  & 45.37  & 51.35  & 46.36  & 51.64  & 42.98  & 55.97  \\
          & 0.7   & 30.73  & 11.68  & 24.83  & 22.08  & 31.62  & 30.04  & 29.39  & 28.95  & 24.95  & 17.55  & 28.65  & 29.71  \\
          & 0.8   & 22.43  & 17.00  & 21.91  & 16.87  & 24.79  & 18.36  & 28.57  & 14.17  & 16.87  & 12.95  & 23.34  & 15.85  \\
          & 0.9   & 21.73  & 8.10  & 20.78  & 10.42  & 21.05  & 8.32  & 22.79  & 12.23  & 19.98  & 9.78  & 14.59  & 7.74  \\
    \bottomrule[1.5pt]
    \end{tabular}%
    }
  \label{tab:re_pruning}%
\end{table*}%

\noindent\textbf{Neural Cleanse.} 
For each class label, Neural Cleanse computes the optimal patch pattern to transform clean inputs into that label. It then identifies potential backdoors by detecting labels with unusually small patterns, using the Anomaly Index metric with a benign/backdoor threshold of $\tau$ = 2.
We test Neural Cleanse on the backdoored ResNet50 models across four datasets and report the results in \cref{fig:defense} (c) - (d).
On all four datasets, the Anomaly Index values of the two backdoor models do not exceed 2, which indicates that our method has successfully passed the tests. 

\begin{remark}[Analysis]
Neural Cleanse relies on the assumption that a fixed class label is disproportionately activated by a minimal trigger, enabling reverse optimization to recover it. However, DarkHash targets image retrieval rather than classification, aiming to shift poisoned samples toward the centroid of a shadow target class in the hash space. By enforcing topological consistency, it creates a structural pull that blends poisoned features into the target class manifold. This violates Neural Cleanse’s core assumption, making it difficult to isolate a minimal trigger or identify any anomalous class—thereby allowing DarkHash to evade detection.
\end{remark}


\noindent\textbf{STRIP.} 
We use STRIP with its default experimental settings to detect our backdoored ResNet50 model on the Pascal VOC dataset, and utilize the \textit{False Alarm Rate} (FAR) used in STRIP to calculate the probability that a backdoored input is recognized as a benign input. As shown in \cref{fig:defense} (e) - (f), the performance and entropy distributions of the backdoored model are similar to those of the benign model, which indicates that STRIP is ineffective against the DarkHash method. 

\begin{remark}[Analysis]
STRIP assumes backdoored inputs produce low-entropy predictions under perturbation. Yet, DarkHash operates in retrieval settings without deterministic labels, leading to uniformly high entropy for both clean and poisoned samples—thereby evading STRIP.
\end{remark}

\noindent\textbf{SentiNet.}  
This method employs Grad-CAM~\cite{selvaraju2017grad} to generate heatmap s that highlight model focus areas, and then occludes or removes these regions to detect potential backdoor triggers affecting classification. 
We apply SentiNet to a ResNet50 backdoored model on the Pascal VOC dataset to evaluate its ability to accurately identify triggers attached to benign samples. 
We measure the overlap between the Grad-CAM region and the trigger, calculating the proportion \(p\) of this overlap relative to the total trigger area. If \(p\) exceeds a threshold \(\theta\), SentiNet detects the trigger. Applying SentiNet to 1000 poisoned samples, we report detection rates for \(\theta\) values of 0.3, 0.4, 0.5, and 0.6 in \cref{fig:defense} (g) - (h).  
We also generate heatmap s that highlight the backdoored model's focus areas and then overlay these heatmap s onto the original images for different triggers. As shown in  \cref{fig:sentinet}, the results indicate that this approach often fails to accurately detect the trigger regions.
The results show that SentiNet often fails to detect the trigger regions.

\begin{remark}[Analysis]
SentiNet detects backdoors by identifying localized trigger regions via saliency mAP s. However, DarkHash does not rely on spatially localized activations. Instead, its topological loss globally aligns poisoned inputs with the target class in the hash space, distributing the trigger’s influence across the representation. This diffuse effect reduces saliency, preventing SentiNet from isolating or neutralizing the backdoor region—thus allowing DarkHash to evade detection.
\end{remark}

\noindent\textbf{Adaptive Defense.} 
We consider a more challenging setting where the defender knows the location of the backdoor injection layer. In our threat model, end users directly adopt pretrained models without further fine-tuning or architectural modifications. Therefore, removing layers (e.g., the last or penultimate layer) is typically impractical, as it alters intermediate feature dimensions and risks breaking the model. To evaluate the potential of mitigating backdoor effects, we simulate adaptive defenses by pruning a large portion of model parameters (from $50\%$ to $90\%$). We conduct experiments on CSQ and HashNet with three backbones (64-bit hash codes), using ImageNet as the surrogate dataset and Pascal VOC as the main task.
As shown in \cref{tab:re_pruning}, when the pruning ratio increases and the t-mAP  significantly drops, the clean mAP  also sharply degrades. This indicates that such aggressive pruning fails to defend against our backdoor attack while severely harming the model’s performance.

\vspace{0.2cm}

\section{Limitations}
In this work, we primarily focus on backdoor threats targeting deep hashing models for image retrieval. 
Due to structural differences between hashing-based models and models used in classification or object detection, we have not extended our approach to those tasks. 
However, we believe that our proposed backdoor attack loss and topological alignment loss introduce a novel and generalizable perspective for backdoor research. Future work can explore how to adapt these components to other tasks beyond retrieval.
Moreover, our work lacks a theoretical analysis of why the proposed attack bypasses existing defenses. As no defense methods are specifically designed for deep hashing models, we evaluate against classification-based defenses. Our attack relies on heuristic design, and the observed effectiveness is supported empirically rather than theoretically. Developing dedicated defenses and theoretical foundations for backdoor vulnerabilities in retrieval models remains future work.

\section{Conclusion}
In this paper, we propose DarkHash, the first data-free backdoor attack against deep hashing models. Specifically, we design a brand-new shadow backdoor attack framework guided by dual semantics. This framework embeds backdoor functionality and preserves the original retrieval accuracy by fine-tuning only specific layers of the victim model using a surrogate dataset.
Experimental results on four image datasets, five model architectures, and two hashing methods demonstrate that DarkHash is highly effective, outperforming existing SOTA backdoor attack methods for deep hashing. Defense experiments show that DarkHash can withstand current mainstream backdoor defense strategies.

\section*{Acknowledgements}
Shengshan Hu's work is supported by the National Natural Science Foundation of China under Grant No.62372196.
Minghui Li's work is supported by the National Natural Science Foundation of China under Grant No. 62202186. 

\bibliographystyle{IEEEtran}
\bibliography{ref}

\end{document}